%% file: acl_latex.tex
\documentclass[11pt]{article}

\usepackage[final]{acl}

\usepackage{times}
\usepackage{latexsym}

\usepackage{booktabs} 
\usepackage{multirow} 
\usepackage{algorithm} 
\usepackage{algorithmic}
\usepackage{colortbl}
\usepackage{xcolor}
\definecolor{Gray}{gray}{0.95}
\usepackage{tcolorbox}
\usepackage{enumitem}
\tcbuselibrary{skins, breakable, theorems} 
\usepackage[T1]{fontenc}

\usepackage[utf8]{inputenc}

\usepackage{microtype}

\usepackage{inconsolata}

\usepackage{graphicx}

\usepackage{amsmath}
\usepackage{amssymb}
\usepackage{xspace}

\usepackage{bm} 
\newcommand{\our}{PILOT\xspace}
\definecolor{OursBlue}{rgb}{0.92, 0.96, 1.0}
\definecolor{HeaderGray}{gray}{0.95}

\title{PILOT: Planning via Internalized Latent Optimization Trajectories for Large Language Models}

\author{
  Haoyu Zheng\textsuperscript{1},
  Yun Zhu\textsuperscript{2},
  Yuqian Yuan\textsuperscript{1},
  Bo Yuan\textsuperscript{1}, \\
  Wenqiao Zhang\textsuperscript{1}\thanks{Corresponding author.},
  Siliang Tang\textsuperscript{1},
  Jun Xiao\textsuperscript{1} \\
  \textsuperscript{1}Zhejiang University \\
  \textsuperscript{2}Shanghai AI Laboratory
}

\begin{document}
\maketitle

\newtcolorbox{promptbox}[1][]{%
  enhanced,
  colback=gray!12,        
  colframe=black!30,      
  boxrule=0.5pt,
  arc=2mm,
  left=8pt,right=8pt,top=6pt,bottom=6pt,
  title=#1,
  fonttitle=\bfseries,    
  coltitle=black,         
  colbacktitle=gray!12,   
  titlerule=0pt,        
  boxed title style={frame hidden},
  before skip=6pt,
  after skip=6pt,
  before upper={\setlength{\itemsep}{0pt}\setlength{\parsep}{0pt}}  
}

\input{sec/0abstract}

\input{sec/1introduction}
\input{sec/2relatedworks}
\input{sec/3method}
\input{sec/4experiments}
\input{sec/5conclusion}

\section{Limitations}

While \textbf{\our} offers a robust alternative to search-based reasoning with zero recurrent decoding latency, we acknowledge specific limitations regarding deployment complexity and generalization:

\paragraph{Data Construction Overhead.}
A core component of our framework is the \textit{Construct-and-Verify} pipeline, which distills expert guidance into high-fidelity \textbf{latent anchors}. While this process is crucial for performance, it introduces additional preprocessing complexity compared to standard Supervised Fine-Tuning on raw datasets. The requirement to synthesize and filter high-quality trajectories creates a trade-off where we accept higher offline data preparation costs to achieve maximum efficiency during online inference.

\paragraph{Domain-Specific Hyperparameters.}
Our analysis highlights that the optimal \textbf{anchoring depth} (the pivot layer) shifts depending on the task nature—deeper for abstract mathematics and shallower for structural code generation. Currently, this insertion layer is treated as a static hyperparameter per domain. Although effective, this requires empirical tuning when adapting the framework to new domains, and a fully dynamic, instance-wise layer selection mechanism remains a direction for future work.

\bibliography{custom}

\newpage

\appendix
\input{sec/6appendix}

\end{document}

%% file: sec/0abstract.tex
\begin{abstract}
Strategic planning is critical for multi-step reasoning, yet compact Language Language Models (LLMs) often lack the capacity to formulate global strategies, leading to error propagation in long-horizon tasks. 
Our analysis reveals that LLMs possess latent reasoning capabilities that can be unlocked when conditioned on explicit plans from a teacher model; however, runtime reliance on external guidance is often impractical due to latency and availability constraints. To bridge this gap, we propose \textbf{PILOT} (\textbf{P}lanning via \textbf{I}nternalized \textbf{L}atent \textbf{O}ptimization \textbf{T}rajectories), a non-invasive framework designed to internalize the strategic oversight of large models into intrinsic \emph{Latent Guidance}. Instead of altering backbone weights, PILOT employs a lightweight Hyper-Network to synthesize a query-conditioned \emph{Latent Guidance}. This vector acts as an internal steering mechanism, guiding the model’s representations toward optimal reasoning paths. Extensive experiments on mathematical and coding benchmarks demonstrate that PILOT effectively stabilizes reasoning trajectories, consistently outperforming strong baselines (e.g., +8.9\% on MATH500) with negligible inference latency. Our code is available at: \url{https://github.com/Chihaya-Anon-chan/PILOT}
\end{abstract}

%% file: sec/1introduction.tex
\section{Introduction}\label{sec:introduction}

Large Language Models (LLMs) have demonstrated remarkable capabilities in complex reasoning, largely driven by the Chain-of-Thought (CoT) paradigm \citep{wei2022chain}. By decomposing problems into intermediate steps, CoT enables models to tackle tasks previously out of reach. However, reliable multi-step reasoning fundamentally relies on \textbf{Strategic Planning} to maintain global coherence across long-horizon trajectories. While compact models often possess the requisite domain knowledge, they frequently struggle with this strategic oversight—the ability to formulate a high-level approach before execution. Without a global strategy, these models are prone to ``myopic'' generation, where minor errors in early reasoning steps cascade into significant deviations, a phenomenon known as error propagation.

Current approaches to mitigating these failures typically rely on \textbf{external scaffolding}. Techniques such as CoT prompting encourage decomposition but do not inherently instill a global planning mechanism. More advanced strategies employ ``Teacher-Student'' paradigms, where a larger model acts as an external guide to correct the smaller model's trajectory during inference. While effective, this reliance on runtime external guidance is practically prohibitive: it introduces severe latency penalties, increases computational costs, and creates a strict dependency on the availability of superior models. 


To address these latency constraints, recent work has explored internal adaptation. Yet existing methods struggle to preserve the balance between improving reasoning and maintaining broad general capability. \textbf{Static intervention} methods, such as LoRA~\citep{hu2022lora} and ReFT~\citep{wu2024reft}, perform static parameter-efficient adaptation: the learned changes are fixed after training and provide {no explicit, per-instance strategic steering} at inference time, often biasing the model toward brittle reasoning templates that fail on complex, heterogeneous instances. Similarly, \textbf{activation steering} techniques~\citep{panickssery2024steeringllama2contrastive} typically rely on fixed steering vectors, which cannot adapt to the diverse logical demands across queries~\citep{venhoff2025understanding}. Conversely, approaches that attempt to internalize reasoning directly into latent representations, such as Coconut~\citep{hao2024training}, often require invasive training that can disrupt the model’s native representation manifold and induce catastrophic forgetting of pre-trained knowledge~\citep{kirkpatrick2017overcoming, luo2025empirical}. Furthermore, \textbf{guidance-based} strategies such as Soft CoT~\citep{xu2025softcot} can suffer from distribution mismatch between the assistant and the backbone, leading to embedding misalignment that limits effectiveness. Even compute-expanding approaches like {Pause Tokens}~\citep{goyal2023think} provide additional thinking budget but still lack a {strategic anchor} to stabilize long-horizon reasoning.
Consequently, current literature lacks a method capable of internalizing guidance without compromising general capability, leaving a gap for a robust, planning-centric approach.

To bridge this gap, we propose \textbf{PILOT} (\textbf{P}lanning via \textbf{I}nternalized \textbf{L}atent \textbf{O}ptimization \textbf{T}rajectories), a non-invasive framework designed to internalize the strategic oversight of large language models into intrinsic \emph{Latent Guidance}. Rather than relying on runtime external calls or altering backbone weights, PILOT employs a lightweight {Hyper-Network} \citep{ha2016hypernetworks} to dynamically synthesize a query-conditioned guidance vector. This vector acts as an internal steering mechanism, effectively replicating the stabilizing effect of a high-level plan within the model's deep semantic layers \citep{su2025token}. By synthesizing these signals strictly from the input query, PILOT ensures that the intervention is tailored to the specific logical requirements of each instance, guiding the model toward optimal reasoning paths without incurring retrieval latency.


Our contributions are summarized as follows:

\begin{itemize}[leftmargin=15pt]
    \item \textbf{Internalized Planning Paradigm.} We propose moving beyond static tuning to directly stabilize reasoning trajectories via intrinsic \emph{Latent Guidance}, effectively internalizing the strategic foresight of larger models.
    \item \textbf{The PILOT Framework.} We introduce a novel architecture employing a {Hyper-Network} to synthesize query-conditioned guidance vectors, acting as a non-invasive internal steering mechanism to prime the model for complex reasoning.
    \item \textbf{Empirical Effectiveness.} Extensive experiments on mathematical and coding benchmarks demonstrate that PILOT consistently enhances the reasoning quality of compact LLMs (e.g., up to \textbf{+8.9\%} gain on MATH500). Crucially, these gains are achieved with \textbf{near-zero extra latency}, stabilizing \textbf{single-path} reasoning trajectories.
\end{itemize}

%% file: sec/2relatedworks.tex
\section{Related Work}
\label{sec:relatedworks}

\subsection{Evolution of Chain-of-Thought and Verification Strategies}
Chain-of-Thought (CoT) prompting \citep{wei2022chain} has revolutionized LLM reasoning by decomposing problems into intermediate steps, yet its autoregressive nature remains susceptible to error propagation, where minor early deviations lead to cascading hallucinations \citep{dziri2023faith, turpin2023language}. To mitigate this, decoding-level strategies like \textit{Self-Consistency} \citep{wang2022self} and \textit{Tree of Thoughts} \citep{yao2023tree} introduce post-hoc verification by sampling multiple trajectories or performing tree search. Other methods explore reasoning rectification via backward verification \citep{xue2023rcot} or adaptive self-correction \cite{wu2024get, zhang2025ascot}. While effective, these methods treat the model as a black box and incur massive computational overhead---often $10\times$ to $50\times$ the original inference cost---making them impractical for low-latency applications. More importantly, they mask errors through external aggregation rather than fundamentally stabilizing the model's internal reasoning process.

\subsection{Parameter-Efficient Adaptation and Latent Reasoning}
PEFT methods like \textit{LoRA} \citep{hu2022lora} and \textit{Prefix-Tuning} \citep{li2021prefix} adapt models via static updates; however, their instance-agnostic nature lacks the granularity for the query-specific strategic guidance required by complex tasks \cite{sun2025transformer, choi2025teaching}. 
Paradigms for latent reasoning \citep{zhu2025survey}, such as \textit{Quiet-STaR} \citep{zelikman2024quiet} and \textit{Coconut} \citep{hao2024training}, attempt to bypass discrete bottlenecks by moving computation into the hidden space. Yet, these methods often require invasive training that disrupts the model's native manifold, risking catastrophic forgetting of general knowledge. 
Similarly, \textit{Pause Tokens} \citep{goyal2023think} expand inference budgets but provide no strategic anchor against semantic drift. Finally, while \textit{Soft CoT} \citep{xu2025softcot} introduces thought vectors from auxiliary models, it often suffers from distributional mismatches that hinder cross-model alignment.

\subsection{Representation Engineering and Activation Steering}
A more recent paradigm, \textit{Representation Engineering} (RepE) \citep{zou2023representation}, aims to control behavior by editing model activations. Techniques like \textit{ReFT} \citep{wu2024reft} learn low-rank interventions on hidden states, while \textit{Contrastive Activation Addition} (CAA) \citep{panickssery2024steeringllama2contrastive} extracts steering vectors by averaging activation differences from contrastive prompt pairs and injects them during inference. \textit{Prototype-Based Steering} \citep{kayan2025prototype} retrieves task-specific exemplars at inference time to guide generation. However, these methods \cite{turner2023steering, zhao2025steering, tang2025unlocking} typically rely on static intervention vectors or heuristic retrieval from external databases, which may not generalize to novel or diverse queries. Furthermore, naive activation editing often causes ``embedding shock''---a distributional shift that disrupts the model's feature space and degrades stability \citep{zhou2025geometry}. 

Our work, \textbf{\our}, addresses these limitations by employing a \textit{Hyper-Network} to dynamically synthesize query-specific latent anchors. Unlike static PEFT or invasive latent reasoning, \our provides instance-level adaptivity without modifying backbone weights or incurring recurrent costs. By incorporating \textit{Energy-Aligned Injection}, \our ensures manifold consistency, offering a non-invasive and scalable solution for stabilizing single-path reasoning trajectories.

%% file: sec/3method.tex
\section{Preliminaries}
\label{sec:preliminaries}

\subsection{Notations}
\label{sec:notations}

We denote the input query by $x$ and the target output by $y=[r; a]$, where $r$ is the rationale and $a$ is the final answer. Let $\mathcal{M}_\phi$ be a frozen causal language model with parameters $\phi$. For a given sequence, we denote token-level hidden states at layer $l$ by $\{\mathbf{h}_i^{(l)}\}_{i=1}^{n}$, where $\mathbf{h}_i^{(l)} \in \mathbb{R}^d$ and $d$ is the hidden size.

We use $l^\dagger$ to denote the \emph{pivot layer} where the latent anchor is injected. Let $\mathcal{Q}$ be the index set of query tokens, and $\mathcal{G}$ be the index set of guidance tokens in the verified expert prefix $[x; g_{\text{exp}}]$. We denote the extracted homogeneous target state by $\mathbf{z}^* \in \mathbb{R}^d$ and the predicted anchor by $\hat{\mathbf{z}} \in \mathbb{R}^d$. $\text{LN}(\cdot)$ denotes layer normalization, $\odot$ denotes element-wise multiplication, and $\|\cdot\|_2$ denotes the $\ell_2$ norm.

\subsection{Problem Formulation}
\label{sec:formulation}

Given an input query $x$, the goal is to generate $y=[r; a]$, where $r$ is a rationale and $a$ is the final answer. A causal LLM $\mathcal{M}_\phi$ models $P_\phi(r, a \mid x)$ autoregressively. In standard generation, the decoding trajectory is rigidly determined by the initial latent state induced by $x$. We instead consider a latent-anchored generation process by introducing an anchor vector $\mathbf{z} \in \mathbb{R}^d$ that conditions the autoregressive decoding:
\begin{equation}
    P_\phi(r, a \mid x, \mathbf{z}) = \prod_{t} P_\phi(y_t \mid x, \mathbf{z}, y_{<t})
\end{equation}
Our objective is to learn an anchor adapter $\psi_\theta: x \mapsto \hat{\mathbf{z}}$ that predicts an instance-specific anchor $\hat{\mathbf{z}}$ from $x$, which is then injected at a pivot layer $l^\dagger$ to stabilize the subsequent reasoning steps.

\section{The \our Framework}
\label{sec:method}

\begin{figure*}[t]
    \centering
    \includegraphics[width=\textwidth]{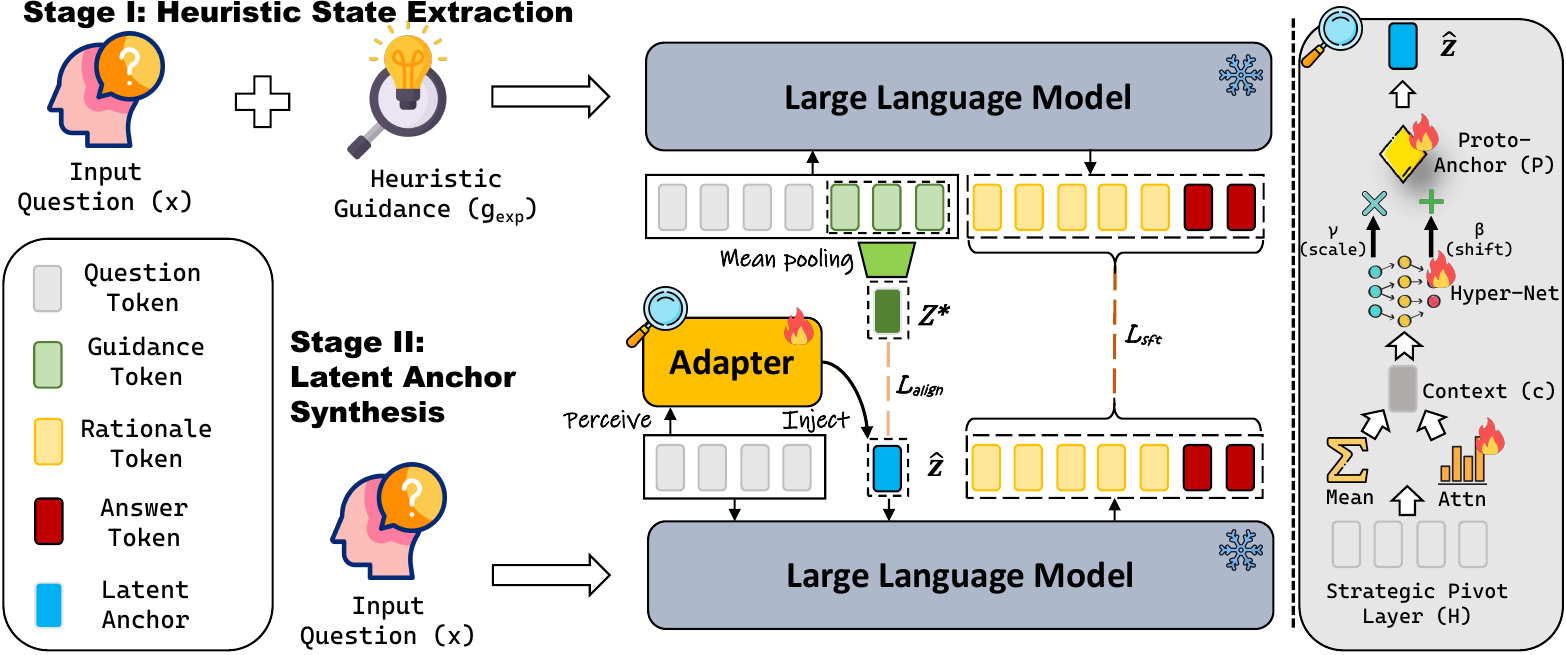}
    \caption{\textbf{The \our Framework Architecture.} \textbf{(Top)} Stage I: Heuristic State Extraction extracting the optimized latent state $\mathbf{z}^*$ from verified expert trajectories. \textbf{(Bottom)} Stage II: Latent Anchor Synthesis during inference predicting $\hat{\mathbf{z}}$ from query tokens. \textbf{(Right)} The Anchor Adapter modulates a \textbf{Proto-Anchor} $\mathbf{P}$ via a Hyper-Network $\mathcal{H}_\theta$ and injects it into the backbone via energy-aligned injection.}
    \label{fig:architecture}
    \vspace{-0.4em}
\end{figure*}

In this section, we build on the formulation in Section~\ref{sec:formulation} and present the key components of \our. We first describe how to extract the homogeneous target state $\mathbf{z}^*$ (Section~\ref{sec:target_construction}), then introduce the anchor adapter (Section~\ref{sec:adapter}) and injection mechanism (Section~\ref{sec:injection}). Finally, we present the optimization objective (Section~\ref{sec:optimization}). An overview of the pipeline is illustrated in Figure~\ref{fig:architecture}.




\subsection{Target State Extraction}
\label{sec:target_construction}
\begin{figure}[t]
    \centering
    \includegraphics[width=\columnwidth]{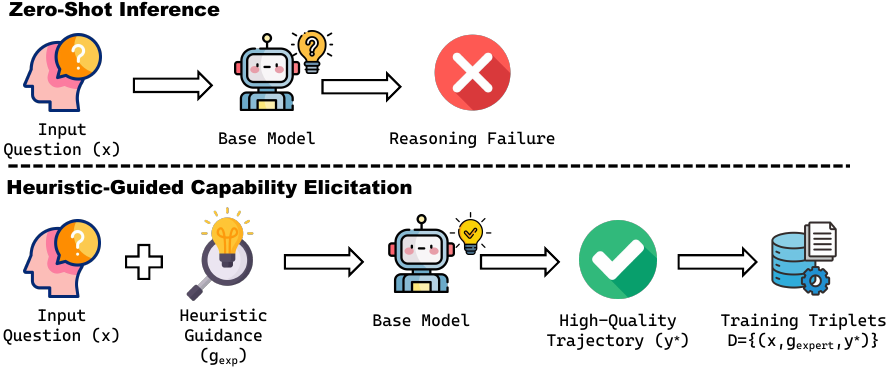}
    \caption{\textbf{Data Construction via Construct-and-Verify.} We filter for hard instances where the base model fails zero-shot but succeeds with expert guidance $g_{\text{exp}}$. These verified triplets $(x, g_{\text{exp}}, y^*)$ form the training set $\mathcal{D}_{\text{train}}$.}
    \label{fig:data_pipeline}
\end{figure}

To obtain high-fidelity supervision signals, we utilize a \textbf{Construct-and-Verify} pipeline (Figure~\ref{fig:data_pipeline}) to derive the \textit{homogeneous target state} $\mathbf{z}^*$.

\paragraph{Verification and Blind-Test.} 
For each query $x$, we identify failure cases of the base model $\mathcal{M}_\phi$. We then generate expert \textit{Heuristic Guidance} ($g_{\text{exp}}$). To ensure $g_{\text{exp}}$ provides strategic anchoring rather than a direct shortcut to the answer, we perform a \textbf{blind test}: if the model solves the problem given $g_{\text{exp}}$ alone (without $x$), the sample is discarded. This ensures $\mathcal{D}_{\text{train}} = \{(x, g_{\text{exp}}, y^*)\}$ captures genuine strategic intent.

\paragraph{Homogeneous Target Projection.} 
To ensure vector-space compatibility, we process the verified sequence $[x; g_{\text{exp}}]$ through the frozen reference model. The homogeneous target state vector $\mathbf{z}^*$ is extracted at the output of the pivot layer $l^\dagger$ by mean-pooling over the \textbf{guidance tokens} $\mathcal{G}$:
\begin{equation}
    \mathbf{z}^* = \frac{1}{|\mathcal{G}|} \sum_{i \in \mathcal{G}} \mathbf{h}_i^{(l^\dagger)}
\end{equation}
Extracting $\mathbf{z}^*$ from $\mathcal{G}$ allows the vector to encapsulate the "optimized" state reached by successful reasoning trajectories, serving as the ground-truth for alignment.

\subsection{Anchor Adapter Architecture}
\label{sec:adapter}

The anchor adapter $\psi_\theta$ serves as a perceiver that synthesizes an anchoring signal while strictly respecting causal constraints.

\paragraph{Dual-Channel Context Aggregation.}
The adapter captures query semantics through a residual fusion of global semantics (via Mean-Pooling) and salient entity features (via Attention-Pooling). Given question features $\mathbf{H}_{\mathcal{Q}}$ (i.e., the backbone hidden states of query tokens at the pivot depth, $\mathbf{H}_{\mathcal{Q}}=\{\mathbf{h}_i^{(l^\dagger)}\}_{i\in\mathcal{Q}}$), the context vector $\mathbf{c}_Q$ is derived as:
\begin{equation}
    \begin{aligned}
    \mathbf{c}_Q ={}& \underbrace{\text{MeanPool}(\mathbf{H}_{\mathcal{Q}})}_{\text{Global Intent}} \\
    &+ \underbrace{\sum_{i \in \mathcal{Q}} \text{softmax}(\mathbf{w}_a^T \mathbf{h}_i^{(l^\dagger)}) \mathbf{h}_i^{(l^\dagger)}}_{\text{Salient Entities}}
    \end{aligned}
\end{equation}
where $\mathbf{w}_a \in \mathbb{R}^d$ is a learnable attention query. This design ensures $\mathbf{c}_Q$ captures both holistic sentence structure and key logical entities.

\paragraph{Proto-Anchor Modulation.}
We introduce a learnable \textbf{Proto-Anchor} vector $\mathbf{P} \in \mathbb{R}^d$ (a.k.a. a proto-thought prior), acting as a global prior of the reasoning manifold. Crucially, rather than random initialization, we \textbf{warm-start} $\mathbf{P}$ with the global centroid of target states $\mathbb{E}[\mathbf{z}^*]$ computed over a subset of training data. 

We employ a \textbf{Hyper-Network} $\mathcal{H}_\theta$ to predict channel-wise FiLM modulation coefficients $[\gamma; \beta]$ from $\mathbf{c}_Q$:
\begin{align}
    [\gamma; \beta] &= \mathcal{H}_\theta(\mathbf{c}_Q) \\
    \mathbf{v}_{raw} &= \gamma \odot \mathbf{P} + \beta
\end{align}
To ensure training stability, $\mathcal{H}_\theta$ is initialized as an \textbf{Identity Prior} (i.e., weights $\approx 0$, bias $\gamma=1, \beta=0$). This forces the optimization to start from the stable global prototype $\mathbf{P}$ and progressively learn instance-specific deviations.

\subsection{Anchor Injection Mechanism}
\label{sec:injection}

\paragraph{Delayed Visibility Masking.} 
To integrate the anchoring signal non-invasively, we append a placeholder token to the query. Concretely, the placeholder is appended to the end of the input context and is not part of the textual output; we assign it a dedicated learnable embedding and intervene only on its hidden state, leaving all other token states unchanged. For all layers $l < l^\dagger$, this token is isolated via a causal mask. At the pivot layer $l^\dagger$, its visibility is enabled, and its hidden state is replaced by $\hat{\mathbf{z}}$.

\paragraph{Energy-Aligned Injection.}
To reconcile directional anchoring with norm-sensitive attention mechanisms, we propose an \textbf{Energy-Aligned Injection} that decouples semantic orientation from physical intensity:
\begin{equation}
    \hat{\mathbf{z}} = \operatorname{Softplus}(\alpha) \cdot \sigma_{\text{ctx}} \cdot \frac{\text{LN}(\mathbf{v}_{\text{raw}})}{\|\text{LN}(\mathbf{v}_{\text{raw}})\|_2}
\end{equation}
where $\sigma_{\text{ctx}} = \operatorname{Mean}_{i \in \mathcal{Q}} \|\mathbf{h}_i^{(l^\dagger)}\|_2$ adapts the injection scale to the current context energy (computed over query tokens, as in our implementation). $\alpha$ is a zero-initialized learnable scalar. To prevent the injection from overwhelming intrinsic backbone features, we apply a regularization penalty if the gating scale $\text{Softplus}(\alpha)$ exceeds a threshold $\tau=2.0$.

\subsection{Optimization Objectives}
\label{sec:optimization}

The framework is optimized via a two-phase curriculum. 

\paragraph{Phase 1: Latent Alignment.} 
We freeze the backbone and minimize the cosine distance between the predicted anchor vector and the homogeneous target: $\mathcal{L}_{\text{align}} = 1 - \cos(\hat{\mathbf{z}}, \mathbf{z}^*)$. This phase grounds the adapter in the expert reasoning manifold.

\paragraph{Phase 2: Anchored Fine-Tuning.} 
We keep the backbone frozen and optimize the adapter components (including the hyper-network, $\mathbf{P}$, and the gate scalar) using the SFT loss $\mathcal{L}_{\text{SFT}}$, while retaining the alignment loss and the gate regularization as structural constraints:
\begin{equation}
    \mathcal{L}_{\text{total}} = \mathcal{L}_{\text{SFT}} + \lambda_1 \mathcal{L}_{\text{align}} + \lambda_2 \mathcal{L}_{\text{gate}}
\end{equation}
where $\mathcal{L}_{\text{gate}} = \max(0, \operatorname{Softplus}(\alpha) - 2.0)^2$. We set $\lambda_1 = 0.1$ and $\lambda_2 = 0.01$, anchoring the signal to the expert manifold while preventing "embedding shock" via norm constraints.

%% file: sec/4experiments.tex
\section{Experiments}
\label{sec:experiments}

\begin{table*}[!t]
\centering
\small
\renewcommand{\arraystretch}{1.1} 

\definecolor{OursBlue}{rgb}{0.92, 0.96, 1.0} 
\definecolor{HeaderGray}{gray}{0.95}

\caption{\textbf{Main Results.} Pass@1 accuracy (Mean $\pm$ Std over 5 runs). \textbf{\our} (blue rows) consistently outperforms baselines across all model scales (1.5B, 7B, 8B), maintaining robustness even on saturated tasks like GSM8K.}
\label{tab:main_results}

\setlength{\tabcolsep}{3.5pt} 
\resizebox{\textwidth}{!}{
\begin{tabular}{l ccc ccc ccc}
\toprule
\multicolumn{10}{c}{\cellcolor{HeaderGray}\textsc{\textbf{Panel A: Mathematical Reasoning}}} \\
\midrule
\multirow{2}{*}{\textbf{Method}} & \multicolumn{3}{c}{\textbf{Qwen2.5-1.5B}} & \multicolumn{3}{c}{\textbf{Qwen2.5-7B}} & \multicolumn{3}{c}{\textbf{Llama-3.1-8B}} \\
\cmidrule(lr){2-4} \cmidrule(lr){5-7} \cmidrule(lr){8-10}
 & \textbf{MATH} & \textbf{AIMO} & \textbf{GSM} & \textbf{MATH} & \textbf{AIMO} & \textbf{GSM} & \textbf{MATH} & \textbf{AIMO} & \textbf{GSM} \\
\midrule
Zero-shot CoT & 43.20 & 22.89 & 66.49 & 71.00 & \underline{43.37} & 88.25 & 47.60 & 20.48 & 83.93 \\
LoRA & $47.24_{\pm 0.67}$ & $\underline{25.54}_{\pm 1.01}$ & $68.32_{\pm 0.72}$ & $72.84_{\pm 0.64}$ & $42.65_{\pm 1.08}$ & $88.22_{\pm 0.58}$ & $47.92_{\pm 1.49}$ & $\underline{22.17}_{\pm 0.66}$ & $\underline{84.22}_{\pm 1.27}$ \\
Soft CoT & $\underline{49.32}_{\pm 0.73}$ & $24.34_{\pm 1.98}$ & $\textbf{69.80}_{\pm 0.81}$ & $\underline{73.20}_{\pm 1.44}$ & $41.93_{\pm 0.54}$ & $\underline{89.05}_{\pm 1.00}$ & $\underline{48.48}_{\pm 0.97}$ & $19.76_{\pm 1.37}$ & $84.20_{\pm 1.25}$ \\
ReFT & $40.20_{\pm 0.92}$ & $20.24_{\pm 1.32}$ & $63.76_{\pm 1.84}$ & $68.88_{\pm 1.03}$ & $40.72_{\pm 0.54}$ & $85.08_{\pm 1.19}$ & $47.08_{\pm 1.08}$ & $20.96_{\pm 1.37}$ & $82.65_{\pm 1.53}$ \\
CAA & 43.80 & 21.69 & 64.75 & 71.60 & 40.96 & 86.28 & 48.20 & 21.69 & 83.17 \\
Pause Token & $43.88_{\pm 1.58}$ & $21.20_{\pm 1.37}$ & $63.56_{\pm 0.32}$ & $69.92_{\pm 1.25}$ & $40.96_{\pm 0.85}$ & $86.08_{\pm 0.29}$ & $47.36_{\pm 1.68}$ & $20.48_{\pm 1.90}$ & $82.68_{\pm 0.41}$ \\
Coconut & $46.36_{\pm 0.64}$ & $23.37_{\pm 1.37}$ & $65.35_{\pm 1.21}$ & $71.72_{\pm 0.76}$ & $42.17_{\pm 0.85}$ & $86.07_{\pm 0.31}$ & $48.08_{\pm 1.32}$ & $18.55_{\pm 0.66}$ & $83.32_{\pm 1.24}$ \\
\rowcolor{OursBlue} \textbf{\our (Ours)} & $\textbf{52.08}_{\pm 0.59}$ & $\textbf{26.27}_{\pm 1.01}$ & $\underline{68.93}_{\pm 1.07}$ & $\textbf{75.24}_{\pm 1.13}$ & $\textbf{45.06}_{\pm 1.37}$ & $\textbf{89.17}_{\pm 0.87}$ & $\textbf{51.64}_{\pm 0.77}$ & $\textbf{24.10}_{\pm 1.90}$ & $\textbf{85.35}_{\pm 0.81}$ \\
\bottomrule
\end{tabular}
}

\vspace{1mm} 

\setlength{\tabcolsep}{10pt} 
\resizebox{\textwidth}{!}{
\begin{tabular}{l cc cc cc}
\toprule
\multicolumn{7}{c}{\cellcolor{HeaderGray}\textsc{\textbf{Panel B: Code Generation}}} \\
\midrule
\multirow{2}{*}{\textbf{Method}} & \multicolumn{2}{c}{\textbf{Qwen2.5-1.5B}} & \multicolumn{2}{c}{\textbf{Qwen2.5-7B}} & \multicolumn{2}{c}{\textbf{Llama-3.1-8B}} \\
\cmidrule(lr){2-3} \cmidrule(lr){4-5} \cmidrule(lr){6-7}
 & \textbf{HEval} & \textbf{MBPP} & \textbf{HEval} & \textbf{MBPP} & \textbf{HEval} & \textbf{MBPP} \\
\midrule
Zero-shot CoT & 46.34 & 42.40 & 71.34 & \underline{57.60} & 53.05 & 53.20 \\
LoRA & $50.12_{\pm 0.80}$ & $\underline{44.24}_{\pm 0.93}$ & $\underline{75.73}_{\pm 0.67}$ & $56.36_{\pm 0.65}$ & $\underline{57.07}_{\pm 0.33}$ & $\underline{54.48}_{\pm 0.67}$ \\
Soft CoT & $\underline{50.49}_{\pm 0.67}$ & $41.16_{\pm 1.63}$ & $71.83_{\pm 1.64}$ & $56.04_{\pm 0.74}$ & $53.17_{\pm 0.67}$ & $53.28_{\pm 0.70}$ \\
ReFT & $42.80_{\pm 1.17}$ & $39.80_{\pm 0.45}$ & $68.66_{\pm 1.47}$ & $54.40_{\pm 0.75}$ & $51.34_{\pm 1.09}$ & $52.04_{\pm 0.70}$ \\
CAA & 43.90 & 40.60 & 69.51 & 55.20 & 51.83 & 52.60 \\
Pause Token & $44.88_{\pm 0.70}$ & $41.88_{\pm 1.24}$ & $68.41_{\pm 0.80}$ & $55.64_{\pm 1.32}$ & $52.07_{\pm 2.05}$ & $52.12_{\pm 0.87}$ \\
Coconut & $47.68_{\pm 1.52}$ & $39.84_{\pm 1.60}$ & $69.39_{\pm 0.51}$ & $54.80_{\pm 1.14}$ & $52.56_{\pm 0.90}$ & $51.32_{\pm 0.97}$ \\
\rowcolor{OursBlue} \textbf{\our (Ours)} & $\textbf{56.34}_{\pm 0.33}$ & $\textbf{46.36}_{\pm 0.48}$ & $\textbf{77.44}_{\pm 1.43}$ & $\textbf{61.60}_{\pm 0.66}$ & $\textbf{58.41}_{\pm 2.04}$ & $\textbf{56.80}_{\pm 0.72}$ \\
\bottomrule
\end{tabular}
}
\end{table*}

\subsection{Experimental Setup}

\paragraph{Datasets \& Filtering.}
We evaluate \our across \textbf{Mathematics} and \textbf{Code Generation}. 
A core component is our \textit{Construct-and-Verify} pipeline, which distills training sets into compact, model-specific subsets ($\mathcal{D}_{\text{train}}$).
As shown in Table~\ref{tab:data_stats}, we use a \textit{Boundary Filter} for {MATH}~\citep{hendrycksmath2021} to capture the reasoning frontier, and a \textit{Refinement Filter} for MBPP~\citep{austin2021program} to align structural logic.
Performance is evaluated on MATH500~\citep{hendrycksmath2021} (primary), AIMO Val. (83 AMC-level problems) \citep{aimo_validation_amc_2024}, and GSM8K (robustness) \citep{cobbe2021gsm8k} for math; and HumanEval \citep{chen2021codex} and MBPP-Test \citep{austin2021program} for coding.

\begin{table}[t]
\centering
\small
\renewcommand{\arraystretch}{1.15} 
\setlength{\tabcolsep}{5pt}        

\caption{\textbf{Data Filtering Statistics.} Retention rate reflects the percentage of samples passing the Construct-and-Verify pipeline. Larger models typically show distinct retention patterns based on task difficulty.}
\label{tab:data_stats}

\resizebox{\columnwidth}{!}{
\begin{tabular}{l c c c r}
\toprule
\textbf{Base Model} & \textbf{Source} & \textbf{Original} & \textbf{Filtered $\mathcal{D}_{train}$} & \textbf{Retention} \\
\midrule

\multicolumn{5}{c}{\cellcolor{gray!15}\textbf{\textsc{Domain: Mathematics}}} \\
\midrule
Qwen2.5-1.5B & \multirow{3}{*}{MATH} & \multirow{3}{*}{7,500} & 1,103 & 14.7\% \\
Qwen2.5-7B   &                       &                        & 543   & 7.2\%  \\
Llama-3.1-8B &                       &                        & 885   & 11.8\% \\

\midrule
\multicolumn{5}{c}{\cellcolor{gray!15}\textbf{\textsc{Domain: Code Generation}}} \\
\midrule
Qwen2.5-1.5B & \multirow{3}{*}{MBPP} & \multirow{3}{*}{374}   & 204   & 54.5\% \\
Qwen2.5-7B   &                       &                        & 267   & 71.4\% \\
Llama-3.1-8B &                       &                        & 244   & 65.2\% \\
\bottomrule
\end{tabular}
}
\end{table}

\paragraph{Base Models.}
We conduct experiments on three instruction-tuned models: Qwen2.5-1.5B-Instruct, Qwen2.5-7B-Instruct \citep{qwen2.5}, and Llama-3.1-8B-Instruct \cite{grattafiori2024llama}. 
For target construction, we utilize DeepSeek-V3.1 as the Expert Model.

\paragraph{Baselines.}
We compare \our against representative methods. To validate \our's ability to improve \textbf{single-path reasoning trajectories}, we select baselines that, like \our, aim to enhance model performance within a single forward pass. For fairness, trainable baselines are fine-tuned using the same filtered subset $\mathcal{D}_{\text{train}}$.
\begin{itemize}
    \item \textbf{Discrete Prompting:} Standard \textbf{Zero-shot CoT}, representing the model's base reasoning capacity without any external intervention.
    \item \textbf{Static Tuning:} Parameter-efficient methods that learn \textbf{fixed adaptation} for the target domain, including \textbf{LoRA} \citep{hu2022lora} and assistant-guided \textbf{Soft CoT} \citep{xu2025softcot}.
    \item \textbf{Latent Intervention:} Methods that directly manipulate internal states, including \textbf{ReFT} \citep{wu2024reft} and \textbf{CAA} \citep{panickssery2024steeringllama2contrastive}. \textbf{CAA} modifies activations via static steering vectors extracted from contrastive reasoning pairs. We also compare against compute-expanding \textbf{Pause Tokens} \citep{goyal2023think} and \textbf{Coconut} \citep{hao2024training} to benchmark against implicit latent reasoning paradigms.
\end{itemize}

\paragraph{Implementation Details.}
Experiments are performed on NVIDIA H20 GPUs. The adapter is trained via a two-stage curriculum: Phase 1 (Alignment) with a learning rate of $1e-4$, and Phase 2 (Anchored SFT) with $2e-5$, using regularization weight $\lambda=0.1$. 
Training epochs are adapted to data scale: we train Math models for 3 epochs per phase, while Coding models undergo prolonged training (10 epochs for Alignment, 8 for SFT) to ensure convergence. 
For evaluation, we employ \textbf{greedy decoding} (temperature=0). To account for training variance, we train 5 independent adapters with different random seeds and report the mean and standard deviation across 5 independent runs.

\subsection{Main Results}
\label{sec:main_results}

Table~\ref{tab:main_results} presents the comprehensive performance. We compare \our against discrete prompting, static tuning, and latent intervention baselines.

\paragraph{Dominance in Complex Reasoning.}
On \textbf{MATH500}, \our consistently outperforms all baselines. Notably, on Qwen2.5-1.5B, \our achieves a remarkable gain, significantly surpassing Soft CoT and LoRA. This confirms dynamic latent anchoring effectively activates dormant reasoning capacity. Even on stronger 7B/8B models, \our maintains a clear edge, suggesting that static tuning struggles to generalize from limited training data (less than 1k samples).

\paragraph{Robustness.}
As seen in Table~\ref{tab:main_results}, several latent-intervention baselines (e.g., \textbf{ReFT} and \textbf{Pause Token}) underperform the base model on GSM8K, suggesting that directly modifying hidden states or introducing unguided extra tokens can be brittle under distribution shift. \textbf{CAA} improves over ReFT and slightly surpasses the Zero-shot baseline on MATH, but does not consistently transfer to GSM8K. \textbf{Coconut} also shows limited gains across settings. In contrast, \our maintains robustness on saturated benchmarks and yields consistent improvements, consistent with our Energy-Aligned Injection being non-invasive.

\paragraph{Code Generalization.}
\our generalizes well on \textbf{HumanEval} (e.g., surpassing LoRA on Qwen-7B). On MBPP, static tuning can sometimes underperform the base model, suggesting potential negative transfer. \our's input-dependent modulation mitigates this effect while improving both coding benchmarks.

\subsection{Additional Validation of Planning Internalization}
\label{sec:additional_validation}

We include three targeted validations to test whether \our's gains can be reproduced by \textit{explicit text-level planning} and whether latent injection harms representation geometry.

\paragraph{Setup of Textual Planning Baselines.}
Reviewers specifically asked whether PILOT's gains could be recovered by making the planning signal explicit in text. We therefore construct two additional baselines on the same 7B backbone. \textbf{External Planning Prompting} simulates ``Stage 1 at inference time'': for each query, we first use \textbf{DeepSeek-V3.1} to generate a concise strategy hint $g_{exp}$ \emph{without access to the ground-truth answer}, and prepend it to the zero-shot prompt. \textbf{LoRA SFT (Text Distillation)} instead fine-tunes the model to explicitly emit heuristic guidance before solving, using the \emph{same} Construct-and-Verify supervision as PILOT ($|\mathcal{D}_{train}|{=}543$ for Math and $267$ for Code). This isolates whether PILOT's benefit comes merely from exposing the heuristic in text, rather than from internalizing it as latent control.

\begin{table*}[t]
\centering
\small
\setlength{\tabcolsep}{10pt}
\caption{\textbf{Latent Planning vs. Explicit Textual Planning.} Pass@1 (\%) on Qwen2.5-7B-Instruct. \textbf{External Planning Prompting} prepends strategy hints generated by a stronger planner at inference time, while \textbf{LoRA SFT (Text Distillation)} trains the model to explicitly output heuristic text before reasoning using the same Construct-and-Verify supervision as \our.}
\label{tab:text_vs_latent}
\resizebox{\textwidth}{!}{
\begin{tabular}{lccccc}
\toprule
\textbf{Method} & \textbf{MATH500} & \textbf{AIMO} & \textbf{GSM8K} & \textbf{HumanEval} & \textbf{MBPP} \\
\midrule
Zero-shot CoT & 71.00 & \underline{43.37} & 88.25 & 71.34 & \underline{57.60} \\
External Planning Prompting & \underline{72.60} & \underline{43.37} & \textbf{89.31} & \underline{73.29} & 54.00 \\
LoRA SFT (Text Distillation) & 72.20 & 42.17 & 87.95 & 70.81 & 55.00 \\
\rowcolor{OursBlue} \textbf{\our (Latent)} & \textbf{75.24} & \textbf{45.06} & \underline{89.17} & \textbf{77.44} & \textbf{61.60} \\
\bottomrule
\end{tabular}
}
\end{table*}

\paragraph{Latent Planning vs. Textual Planning.}
Table~\ref{tab:text_vs_latent} shows a consistent gap between latent and text-level planning. \textbf{External Planning Prompting} improves over Zero-shot CoT on some in-domain settings (e.g., MATH500: $72.60$ vs.\ $71.00$), but still remains clearly below PILOT ($75.24$), and degrades on MBPP ($54.00 < 57.60$). \textbf{LoRA SFT (Text Distillation)} is even less stable: while it learns to verbalize heuristic guidance, it underperforms PILOT on every benchmark and falls below the zero-shot baseline on AIMO and MBPP. These results indicate that simply \emph{showing} or \emph{generating} a plan in natural language is insufficient.

\paragraph{Why Textual Planning Falls Short.}
The failure mode is consistent with the rebuttal hypothesis of a \textbf{text-level information bottleneck}. Once planning is serialized into text, the compact model must first parse the heuristic and then follow it within the same autoregressive stream used for execution. This introduces additional exposure bias and can conflict with the model's native token distribution, especially on out-of-distribution tasks. By contrast, \our internalizes the same strategic information directly in latent space, avoiding this detour through surface-form generation while preserving the low-overhead single-path decoding property.

\begin{table}[t]
\centering
\small
\setlength{\tabcolsep}{7pt}
\caption{\textbf{Representation Isotropy Diagnostics.} Average pairwise cosine similarity of final-layer hidden states over $N{=}100$ correctly solved examples. Lower values indicate more dispersed representations.}
\label{tab:isotropy}
\begin{tabular}{lccc}
\toprule
\textbf{Domain} & \textbf{Base} & \textbf{\our} & \textbf{$\Delta$} \\
\midrule
Math & 0.838 & \textbf{0.692} & -0.146 \\
Code & 0.944 & \textbf{0.867} & -0.077 \\
\bottomrule
\end{tabular}
\end{table}

\paragraph{Isotropy and Oversmoothing.}
To directly test whether latent injection collapses the representation manifold, we extract the final-layer hidden state of the last generated token from $N{=}100$ correctly solved examples on MATH500 and HumanEval, and compute the mean pairwise cosine similarity. Table~\ref{tab:isotropy} shows that the average similarity \emph{decreases} after applying \our in both domains (Math: $0.838 \rightarrow 0.692$; Code: $0.944 \rightarrow 0.867$). In other words, the induced representations become \emph{more} dispersed rather than more collapsed. This directly argues against the oversmoothing hypothesis and supports our design goal that Energy-Aligned Injection should \textbf{steer} the trajectory without compressing the model's native geometry.

\subsection{Ablation Studies}
\label{sec:ablation}

To validate component contributions, we conduct ablation studies across model scales. Table~\ref{tab:ablation} summarizes the results.

\begin{table*}[t]
\centering
\small
\setlength{\tabcolsep}{8pt}
\caption{\textbf{Component Ablation across Scales.} Pass@1 accuracy (Mean $\pm$ Std). \textbf{Math} relies heavily on the \textbf{Hyper-Network} for anchoring, while \textbf{Code} depends on \textbf{Energy-Alignment} to prevent structural collapse. The \textbf{Proto-Thought} prior becomes less critical for Code as model scale increases.}
\label{tab:ablation}
\begin{tabular}{l|cc|cc}
\toprule
\multirow{2}{*}{\textbf{Configuration}} & \multicolumn{2}{c|}{\textbf{Qwen2.5-1.5B}} & \multicolumn{2}{c}{\textbf{Qwen2.5-7B}} \\
 & \textbf{MATH} & \textbf{HEval} & \textbf{MATH} & \textbf{HEval} \\
\midrule
\textbf{\our (Full)} & $\textbf{52.08}_{\pm 0.59}$ & $\textbf{56.34}_{\pm 0.33}$ & $\textbf{75.24}_{\pm 1.13}$ & $\textbf{77.44}_{\pm 1.43}$ \\
\midrule
\textit{Architectural Variants} & & & & \\
(1) w/o Hyper-Net (Static) & $47.72_{\pm 0.99}$ & $50.37_{\pm 1.70}$ & $72.84_{\pm 1.44}$ & $72.68_{\pm 0.67}$ \\
(2) w/o Proto-Thought & $48.84_{\pm 1.13}$ & $\underline{54.39}_{\pm 1.32}$ & $74.52_{\pm 1.40}$ & $\underline{76.71}_{\pm 1.52}$ \\
(3) w/o Energy-Alignment & $\underline{49.16}_{\pm 0.61}$ & $51.22_{\pm 0.43}$ & $\underline{75.04}_{\pm 1.01}$ & $73.17_{\pm 0.61}$ \\
\bottomrule
\end{tabular}
\vspace{-2mm}
\end{table*}

\paragraph{Architectural Ablation.}
Table~\ref{tab:ablation} reveals scale-dependent dynamics.
\textbf{Hyper-Network:} Removing input-dependent anchoring (Row 1) consistently degrades performance across both tasks and scales, confirming that static adapters cannot capture the complexity of reasoning manifolds.
\textbf{Proto-Thought:} While critical for 1.5B, its impact on Code diminishes at 7B ($77.44 \to 76.71$), suggesting larger models may implicitly learn structural priors. However, it remains vital for Math ($75.24 \to 74.52$), where logical planning is less inherent in the pre-training objective.

\begin{figure}[t]
    \centering
    \includegraphics[width=\columnwidth]{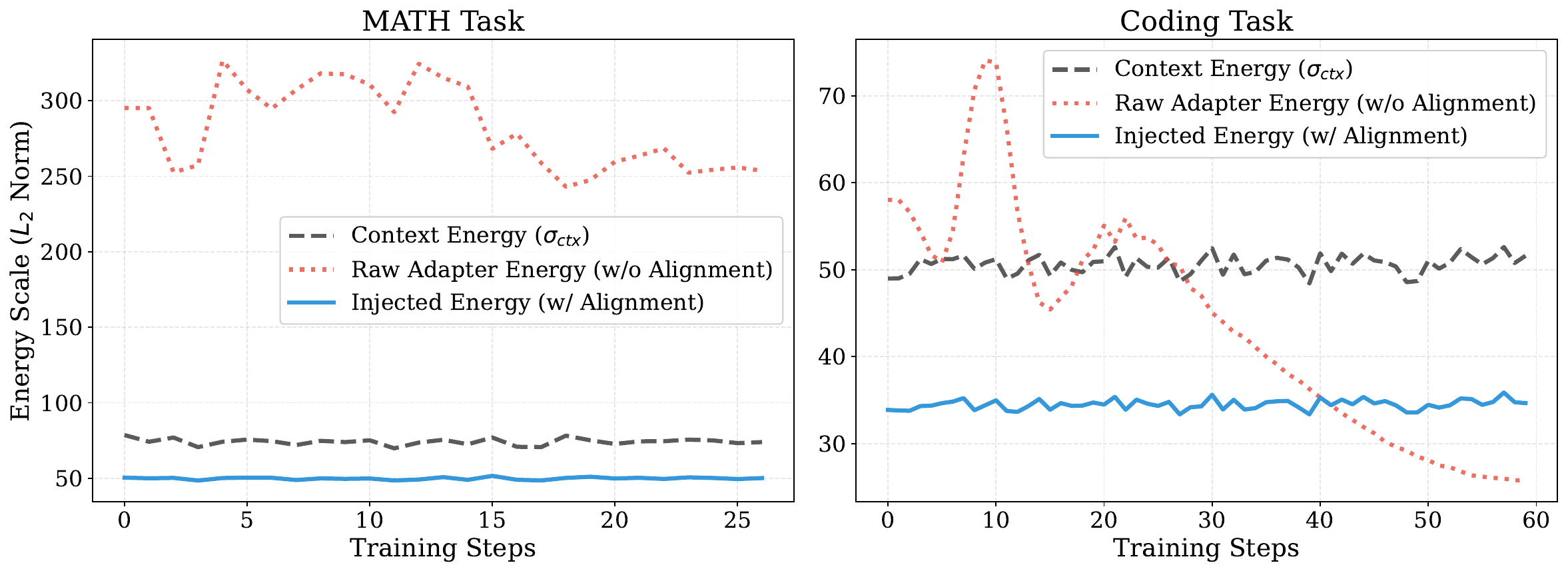}
\caption{\textbf{Energy Alignment Dynamics (7B).} Tracking injection vector $L_2$ norm. \textbf{Left (Math):} Raw energy naturally aligns with context. \textbf{Right (Code):} \our's alignment constrains wild fluctuations, preventing "embedding shock" and ensuring stability.}
    \label{fig:norm_evolution}
\end{figure}

\paragraph{Energy Alignment Analysis.}
A striking divergence appears in Row 3. Removing Energy-Alignment causes a severe drop in Code for 7B ($77.44 \to 73.17$), whereas Math remains unaffected.
To explain this, we visualize the norm evolution in Figure~\ref{fig:norm_evolution}.
In \textbf{Code Generation}, the \textbf{Raw Adapter Energy} (without alignment) exhibits high variance and often exceeds the \textbf{Context Energy} ($\sigma_{ctx}$), risking a "structural shock" that disrupts syntax. \our's alignment mechanism forces the \textbf{Injected Energy} to track $\sigma_{ctx}$, stabilizing the intervention.
In \textbf{Mathematics}, the raw energy naturally converges near the context norm, rendering explicit alignment less critical at this scale. This confirms that coding tasks require stricter "energy preservation" to maintain valid output distributions.

\paragraph{Data Efficiency.}
Separately, we validated our filtering protocol on 1.5B. Training on the full unfiltered dataset degraded performance to 50.85\% (Math) and 52.20\% (Code), confirming the value of signal concentration.

\begin{figure}[t]
    \centering
    \includegraphics[width=\columnwidth]{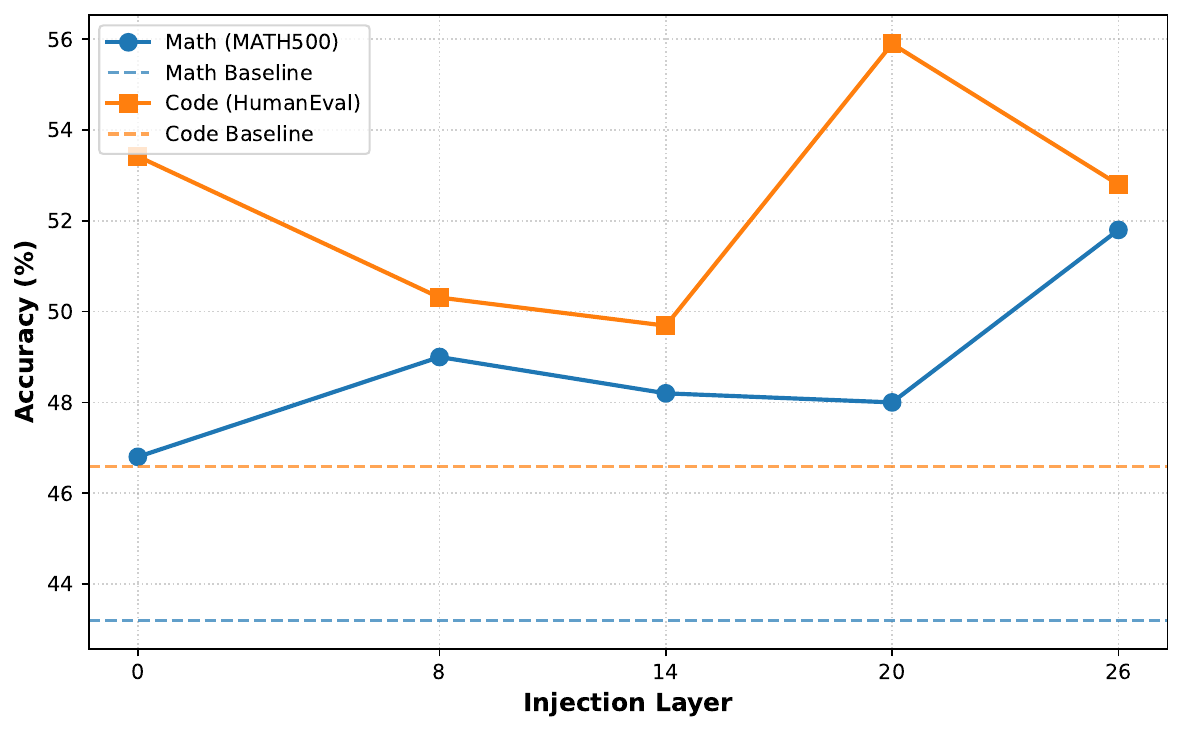}
\caption{\textbf{Injection Depth Sensitivity (Qwen-1.5B).} Optimal pivots shift by task: \textbf{Math} peaks at the deepest layer (26), while \textbf{Code} peaks earlier (20).}
    \label{fig:layer_sensitivity}
\end{figure}

\paragraph{Injection Depth.}
Figure~\ref{fig:layer_sensitivity} shows that \our improves performance across all tested layers. The optimal pivot shifts by task nature: \textbf{Math} peaks at the deepest layer, suggesting abstract reasoning is best guided at the final stage of semantic aggregation. \textbf{Code} peaks earlier, indicating a need to guide structural logic before final syntax rigidifies.

\subsection{Analysis}
\label{sec:analysis}

\paragraph{Efficiency \& Overhead.}
We benchmark inference latency on a single NVIDIA H20 GPU under a unified evaluation protocol (same prompts, decoding settings, and generation budget) across all methods. Table~\ref{tab:efficiency} reports \textbf{prefill (TTFT)} and \textbf{end-to-end} latency averaged over 1k samples.
\textbf{\our} introduces a small prefill overhead of $\mathbf{+3.10}$ \textbf{ms} (21.66 ms vs. 18.56 ms), attributable to a single forward pass of our lightweight Hyper-Network. During decoding, \our injects a static anchor state and therefore does not add per-step recurrent computation, resulting in a negligible total latency increase of $\mathbf{0.2\%}$ (10,230 ms vs. 10,209 ms).
\textbf{Soft CoT} incurs a higher end-to-end latency overhead in our measurement ($\mathbf{+6.2\%}$). We attribute this mainly to its reliance on an additional optimization pipeline for continuous prompts; nevertheless, we follow the official implementation and keep the evaluation protocol consistent across methods.

\begin{figure}[ht]
    \centering
    \includegraphics[width=\columnwidth]{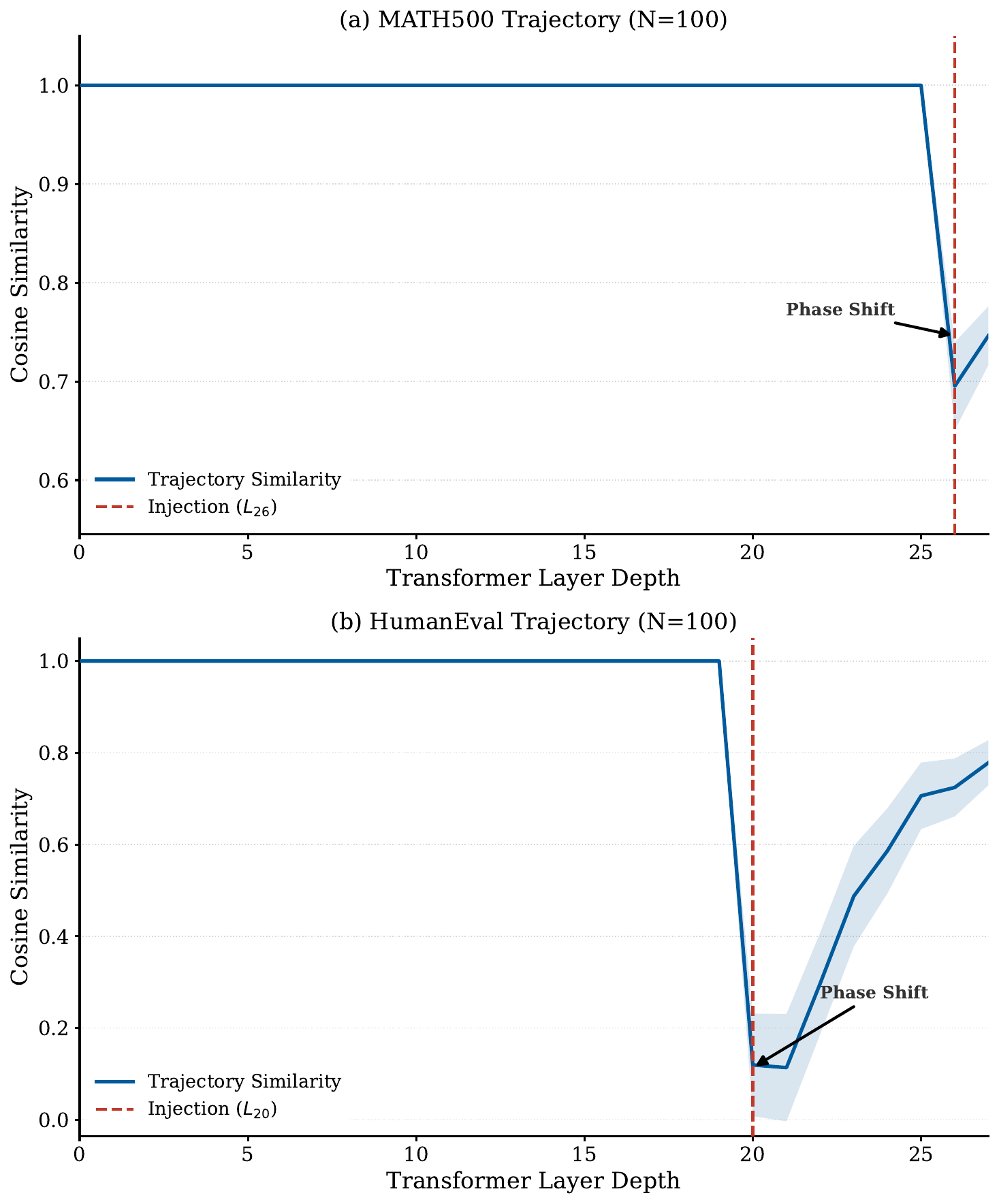}
    \vspace{-6mm}
\caption{\textbf{Mechanistic Visualization of Latent Anchoring.} Cosine similarity between base and anchored states. 
    \textbf{(a) Math:} Layer 26 anchoring indicates "last-mile" correction.
    \textbf{(b) Code:} Layer 20 injection triggers "shock" followed by recovery, implying deep restructuring. Shaded regions: std dev ($N=100$).}
    \label{fig:divergence}
\end{figure}

\begin{table}[h]
\centering
\small
\setlength{\tabcolsep}{4pt}
\caption{\textbf{Inference Latency Analysis.} Benchmarked on Qwen2.5-7B (Avg. over 1k samples). \textbf{\our} incurs negligible decoding overhead compared to \textbf{Soft CoT}.}
\label{tab:efficiency}
\resizebox{1.0\columnwidth}{!}{
\begin{tabular}{l c c c c}
\toprule
\multirow{2}{*}{\textbf{Method}} & \multicolumn{2}{c}{\textbf{Prefill (TTFT)}} & \multicolumn{2}{c}{\textbf{Total Latency}} \\
\cmidrule(lr){2-3} \cmidrule(lr){4-5}
 & \textbf{Time (ms)} & \textbf{Overhead} & \textbf{Time (ms)} & \textbf{Overhead} \\
\midrule
Base Model & 18.56 & -- & 10,209.42 & -- \\
\rowcolor{gray!10} \textbf{\our (Ours)} & \textbf{21.66} & \textbf{+3.10 ms} & \textbf{10,230.52} & \textbf{+0.2\%} \\
Soft CoT & 44.31 & +25.75 ms & 10,842.40 & +6.2\% \\
\bottomrule
\end{tabular}
}
\end{table}

\paragraph{Anchoring Dynamics.}
To understand how \our alters the inference trajectory, we analyze the layer-wise cosine similarity between the base and anchored hidden states (Figure~\ref{fig:divergence}). We observe distinct "Phase Shift" behaviors:
\begin{itemize}
    \item \textbf{Math (Terminal Correction):} For MATH500 (Figure~\ref{fig:divergence}a), anchoring occurs at the deepest semantic pivot (Layer 26). The similarity drops moderately ($\sim 0.7$) and remains divergent. This indicates a \textbf{"last-mile correction"}, where \our refines the final semantic representation right before decoding, ensuring precise logical closure without disrupting the previously accumulated context.
    \item \textbf{Code (Deep Restructuring):} For HumanEval (Figure~\ref{fig:divergence}b), the intervention at Layer 20 triggers a \textbf{massive "injection shock"} (similarity drops to $\sim 0.1$). This implies that for coding, \our fundamentally overwrites the model's internal plan with a new structural blueprint. Crucially, we observe an \textbf{"Assimilation Phase"} (Layers 21-27), where the similarity recovers to $\sim 0.8$. This reveals the model's mechanism: it absorbs the drastic anchoring signal (the drop) and integrates it with its pre-trained linguistic knowledge to generate valid code. The sustained gap ($1.0 \to 0.8$) at the final layer confirms that the output distribution remains successfully shifted.
\end{itemize}

\paragraph{Parameter Efficiency.}
\our is designed for parameter efficiency. For instance, on Qwen2.5-7B (7.6B parameters), it introduces only 38.6M trainable parameters ($\sim 0.5\%$), and on Qwen2.5-1.5B (1.5B parameters), it requires just 7.1M trainable parameters ($\sim 0.46\%$). This lightweight nature allows for rapid adaptation and minimal storage overhead compared to full fine-tuning.

%% file: sec/5conclusion.tex
\section{Conclusion}
\label{sec:conclusion}

In this paper, we study how to stabilize reasoning trajectories in LLMs and propose \textbf{\our} (\textbf{P}lanning via \textbf{I}nternalized \textbf{L}atent \textbf{O}ptimization \textbf{T}rajectories; \textbf{PILOT}). 
\our dynamically primes the model's internal representations with instance-specific anchor vectors, providing a non-invasive way to incorporate heuristic guidance without updating backbone weights. Experiments on challenging mathematics and code-generation benchmarks show consistent gains while introducing minimal decoding overhead in our evaluation setting. Our mechanistic analyses further indicate that the injected anchors can induce a coherent shift in intermediate representations, helping maintain logical consistency over long-horizon generation. 

%% file: sec/6appendix.tex
\section{Instruction Templates}
\label{sec:appendix}
\label{app:prompts}

This section details the instruction templates. We categorize them into \textbf{Task-specific Solvers} (for baseline and verified responses) and \textbf{Heuristic Strategy Generators} (for distilling the expert manifold). To ensure the predicted \textbf{anchor vector} $\hat{\mathbf{z}}$ represents abstract strategic intent, all heuristic templates enforce strict constraints against revealing concrete values or step-by-step solutions.

\subsection{Templates for Mathematical Reasoning (MATH)}

\begin{promptbox}[Template A.1: Math Zero-shot Solver]
You are an AI assistant, you are required to solve mathematical question step by step. Please provide your final answer on a separate line in \verb|\boxed{}| format.
\end{promptbox}

\begin{promptbox}[Template A.2: Heuristic Strategy Generator ($g_{exp}$)]
You are an expert mathematics tutor. Provide a \textbf{strategic insight} to guide problem-solving.
\textbf{CRITICAL RULES:} 
(1) NO numbers, specific values, or final answers. 
(2) NO step-by-step derivations or explicit formulas. 
(3) ONE core strategy in 1-2 sentences.
\textbf{Focus on:} What mathematical concept/method to apply or what key transformation unlocks the problem.
\end{promptbox}

\subsection{Templates for Programming Tasks (MBPP)}

\begin{promptbox}[Template A.5: Programming Zero-shot Solver]
You are a Python programmer. Output ONLY the function implementation. No explanations, no markdown, no comments.
\end{promptbox}

\begin{promptbox}[Template A.6: Programming Heuristic Generator ($g_{exp}$)]
You are an expert programming tutor. Provide a \textbf{strategic insight} to guide problem-solving.
\textbf{CRITICAL RULES:} 
(1) NO code, syntax, or implementation details. 
(2) NO step-by-step logic. 
(3) ONE core strategy in 1-2 sentences.
\textbf{Focus on:} What algorithm or data structure to use.
\end{promptbox}

\newpage

\section{Data Generation Pipeline}
\label{app:pipeline}

The \textbf{Construct-and-Verify} pipeline (as defined in Sec~\ref{sec:target_construction}) identifies high-signal triplets where the \textbf{expert heuristic guidance} $g_{exp}$ serves as a necessary condition for successful reasoning.

\subsection{Pipeline Logic and Pseudocode}

Algorithm~\ref{alg:pipeline} describes the systematic construction of $\mathcal{D}_{train}$. The \textbf{Blind Test} prevents \textbf{anchor vector} contamination by ensuring $g_{exp}$ does not leak the final answer, forcing the adapter to learn strategic anchoring rather than trivial mapping.

\begin{algorithm}[H]
\caption{Construct-and-Verify Pipeline}
\label{alg:pipeline}
\begin{algorithmic}[1]
\REQUIRE Raw dataset $\mathcal{D}_{raw}$, Base model $\mathcal{M}_{\phi}$, Expert model $\mathcal{M}_{exp}$
\ENSURE Filtered dataset $\mathcal{D}_{train}$
\STATE $\mathcal{D}_{train} \leftarrow \emptyset$
\FOR{each question $x \in \mathcal{D}_{raw}$}
    \STATE $y_{zs} \leftarrow \mathcal{M}_{\phi}(\text{Solver}(x))$ \COMMENT{Zero-shot baseline trial}
    \IF{$\text{is\_correct}(y_{zs})$ \AND domain is \texttt{Math}}
        \STATE \textbf{continue} \COMMENT{Boundary Filter: Capture the reasoning frontier}
    \ENDIF
    
    \STATE $g_{exp} \leftarrow \mathcal{M}_{exp}(\text{Heuristic\_Gen}(x))$ \COMMENT{Generate $g_{exp}$ strategic anchor}
    \STATE $y^* \leftarrow \mathcal{M}_{\phi}(\text{Solver}(x, g_{exp}))$ \COMMENT{Verify guidance effectiveness}
    
    \IF{\NOT $\text{is\_correct}(y^*)$}
        \STATE \textbf{continue} \COMMENT{Discard if guidance is insufficient to activate the correct manifold}
    \ENDIF
    
    \STATE \COMMENT{\textbf{Blind Test Stage: Preventing Answer Leakage}}
    \STATE $y_{blind} \leftarrow \mathcal{M}_{\phi}(\text{Solver}(g_{exp}))$ \COMMENT{Check for direct leakage in $g_{exp}$}
    \IF{$\text{is\_correct}(y_{blind})$}
        \STATE \textbf{continue}
    \ENDIF
    
    \STATE $\mathcal{D}_{train} \leftarrow \mathcal{D}_{train} \cup \{(x, g_{exp}, y^*)\}$
\ENDFOR
\RETURN $\mathcal{D}_{train}$
\end{algorithmic}
\end{algorithm}

\subsection{Differentiated Filtering by Domain}
\begin{itemize}
    \item \textbf{Mathematics (Boundary Filtering):} We prioritize instances where the model fails zero-shot. This ensures the \textbf{Anchor Adapter} learns a corrective signal to anchor the hidden states from failure manifolds toward the reasoning manifold defined by $g_{exp}$.
    \item \textbf{Programming (Refinement Filtering):} We retain correct cases to anchor the model toward more \textit{optimized} algorithmic structures, leveraging \our's ability to reinforce efficient coding trajectories.
\end{itemize}

\section{Teacher Capability Sensitivity Analysis}
\label{sec:teacher_impact}

A critical validation of \our is whether the performance gains stem from the \textbf{Energy-Aligned Injection} mechanism itself or merely from distillation of teacher knowledge. We decouple these via a teacher-sensitivity analysis.

\begin{itemize}
    \item \textbf{\our (Self):} The teacher is the base model itself. To derive high-quality \textbf{homogeneous target states} $\mathbf{z}^*$ from the 1.5B model, we utilize a higher sampling budget (32 retries) during the \textbf{Construct} phase to find a successful path.
    \item \textbf{\our (Strong):} The teacher is \textbf{DeepSeek-V3}, providing expert cross-model heuristics to guide the target extraction.
\end{itemize}

As shown in Table~\ref{tab:teacher_ablation}, \textbf{\our (Self)} achieves substantial gains. Since the knowledge source is internal to the model, these results empirically prove that \our's \textbf{Anchor Adapter} effectively stabilizes the model's intrinsic reasoning potential by mitigating autoregressive drift.

\begin{table}[ht]
\centering
\small
\renewcommand{\arraystretch}{1.2}
\caption{\textbf{Teacher Capability Sensitivity Analysis.} Comparison of performance using the model itself (\textit{Self}) vs. \textbf{DeepSeek-V3.1} (\textit{Strong}) as the expert for \textbf{target construction}. Pass@1 (\%). \textit{Base} refers to zero-shot CoT.}
\label{tab:teacher_ablation}
\resizebox{\columnwidth}{!}{
\begin{tabular}{ll ccc c}
\toprule
\textbf{Base Model} & \textbf{Benchmark} & \textbf{Base} & \textbf{\our (Self)} & \textbf{\our (Strong)} & \textbf{$\Delta_{S-S}$} \\ 
\midrule
\multirow{2}{*}{Qwen2.5-1.5B} & MATH500 & 43.20 & 51.80 & \textbf{52.00} & +0.20 \\
                              & Human-Eval & 46.34 & 54.88 & \textbf{56.10} & +1.22 \\ 
\midrule
\multirow{2}{*}{Qwen2.5-7B}   & MATH500 & 71.00 & 73.60 & \textbf{75.20} & +1.60 \\
                              & Human-Eval & 71.34 & \textbf{78.66} & 77.44 & -1.22 \\ 
\bottomrule
\end{tabular}
}
\end{table}

\section{Impact of Training Data Scale and Distribution}
\label{sec:data_scale_impact}

In our main experiments (Sec.~\ref{sec:experiments}), we utilized a filtered subset $\mathcal{D}_{train}$ constructed via our \textit{Construct-and-Verify} pipeline. A natural question arises: \textit{Does reducing the data scale limit the model's potential, and would training on the full official datasets yield better results?}

To address this, we conducted a comprehensive comparative study across all baseline methods, including \textbf{LoRA}, \textbf{Soft CoT}, \textbf{ReFT}, and \textbf{Coconut}. We trained these baselines using two different datasets: the full official training sets ($\mathcal{D}_{full}$) of \textbf{MATH} (7,500 samples) and \textbf{MBPP} (374 samples), and our filtered subset $\mathcal{D}_{train}$. We evaluated performance on \textbf{MATH500} and \textbf{HumanEval} to assess generalization. We also include \textbf{Zero-shot CoT} performance as a reference for the base model's capability.

Table~\ref{tab:data_scale_comparison} presents the comparative results. We observe distinct trends across domains. On \textbf{MATH500}, training on the full dataset ($\mathcal{D}_{full}$) often leads to performance degradation compared to the zero-shot baseline for methods like LoRA, ReFT, and Coconut. This suggests that the distribution mismatch between the ground-truth solutions in $\mathcal{D}_{full}$ and the model's internal reasoning manifold causes catastrophic forgetting. However, \textbf{Soft CoT} is a notable exception, maintaining robustness even with full data. In contrast, our filtered subset $\mathcal{D}_{train}$ consistently outperforms $\mathcal{D}_{full}$ and generally improves upon the zero-shot baseline, highlighting the importance of on-manifold data. On \textbf{HumanEval}, $\mathcal{D}_{train}$ consistently outperforms or matches $\mathcal{D}_{full}$, further validating our approach.

\begin{table*}[h]
\centering
\small
\renewcommand{\arraystretch}{1.15}
\caption{\textbf{Impact of Training Data Scale on Baselines.} Comparison of performance when trained on the full official dataset ($\mathcal{D}_{full}$) versus our filtered, model-aligned subset ($\mathcal{D}_{train}$). \textbf{Zero-shot CoT} represents the base model performance without fine-tuning. On MATH500, training on $\mathcal{D}_{full}$ often hurts performance (falling below Zero-shot CoT) due to distribution shift, whereas $\mathcal{D}_{train}$ yields consistent gains. Soft CoT is an exception, benefiting from $\mathcal{D}_{full}$ but still performing best with $\mathcal{D}_{train}$. All results are reported as Mean $\pm$ Std over 5 runs (except for CAA, which is deterministic).}
\label{tab:data_scale_comparison}
\resizebox{\textwidth}{!}{
\begin{tabular}{ll cc c cc c}
\toprule
\multirow{2}{*}{\textbf{Base Model}} & \multirow{2}{*}{\textbf{Method}} & \multicolumn{3}{c}{\textbf{MATH500}} & \multicolumn{3}{c}{\textbf{HumanEval}} \\
\cmidrule(lr){3-5} \cmidrule(lr){6-8}
 & & \textbf{Full ($\mathcal{D}_{full}$)} & \textbf{Filtered ($\mathcal{D}_{train}$)} & \textbf{$\Delta$} & \textbf{Full ($\mathcal{D}_{full}$)} & \textbf{Filtered ($\mathcal{D}_{train}$)} & \textbf{$\Delta$} \\ 
\midrule
\multirow{6}{*}{Qwen2.5-1.5B} 
 & \textit{Zero-shot CoT} & \multicolumn{2}{c}{43.20} & -- & \multicolumn{2}{c}{46.34} & -- \\
 & LoRA     & 42.88$_{\pm 1.45}$ & \textbf{47.24}$_{\pm 0.67}$ & +4.36 & 48.90$_{\pm 1.00}$ & \textbf{50.12}$_{\pm 0.80}$ & +1.22 \\
 & Soft CoT & 45.20$_{\pm 0.87}$ & \textbf{49.32}$_{\pm 0.73}$ & +4.12 & 50.24$_{\pm 1.02}$ & \textbf{50.49}$_{\pm 0.67}$ & +0.24 \\
 & ReFT     & 40.12$_{\pm 0.79}$ & \textbf{40.20}$_{\pm 0.92}$ & +0.08 & 42.68$_{\pm 0.96}$ & \textbf{42.80}$_{\pm 1.17}$ & +0.12 \\
 & CAA      & 42.20 & \textbf{43.80} & +1.60 & 42.68 & \textbf{43.90} & +1.22 \\
 & Pause Token & 42.32$_{\pm 0.48}$ & \textbf{43.88}$_{\pm 1.58}$ & +1.56 & 43.78$_{\pm 1.00}$ & \textbf{44.88}$_{\pm 0.70}$ & +1.10 \\
 & Coconut  & 41.08$_{\pm 1.68}$ & \textbf{46.36}$_{\pm 0.64}$ & +5.28 & 47.07$_{\pm 1.09}$ & \textbf{47.68}$_{\pm 1.52}$ & +0.61 \\
\midrule
\multirow{6}{*}{Qwen2.5-7B}   
 & \textit{Zero-shot CoT} & \multicolumn{2}{c}{71.00} & -- & \multicolumn{2}{c}{71.34} & -- \\
 & LoRA     & 70.56$_{\pm 1.04}$ & \textbf{72.84}$_{\pm 0.64}$ & +2.28 & 74.39$_{\pm 0.96}$ & \textbf{75.73}$_{\pm 0.67}$ & +1.34 \\
 & Soft CoT & 71.60$_{\pm 1.20}$ & \textbf{73.20}$_{\pm 1.44}$ & +1.60 & 71.10$_{\pm 1.91}$ & \textbf{71.83}$_{\pm 1.64}$ & +0.73 \\
 & ReFT     & 67.24$_{\pm 0.33}$ & \textbf{68.88}$_{\pm 1.03}$ & +1.64 & 68.05$_{\pm 0.55}$ & \textbf{68.66}$_{\pm 1.47}$ & +0.61 \\
 & CAA      & 69.80 & \textbf{71.60} & +1.80 & 68.29 & \textbf{69.51} & +1.22 \\
 & Pause Token & 70.08$_{\pm 1.44}$ & \textbf{69.92}$_{\pm 1.25}$ & -0.16 & 69.15$_{\pm 1.65}$ & \textbf{68.41}$_{\pm 0.80}$ & -0.73 \\
 & Coconut  & 69.24$_{\pm 0.61}$ & \textbf{71.72}$_{\pm 0.76}$ & +2.48 & 67.80$_{\pm 0.90}$ & \textbf{69.39}$_{\pm 0.51}$ & +1.59 \\
\midrule
\multirow{6}{*}{Llama-3.1-8B} 
 & \textit{Zero-shot CoT} & \multicolumn{2}{c}{47.60} & -- & \multicolumn{2}{c}{53.05} & -- \\
 & LoRA     & 47.36$_{\pm 1.55}$ & \textbf{47.92}$_{\pm 1.49}$ & +0.56 & 56.22$_{\pm 0.90}$ & \textbf{57.07}$_{\pm 0.33}$ & +0.85 \\
 & Soft CoT & 47.36$_{\pm 0.86}$ & \textbf{48.48}$_{\pm 0.97}$ & +1.12 & 52.20$_{\pm 1.53}$ & \textbf{53.17}$_{\pm 0.67}$ & +0.98 \\
 & ReFT     & 45.48$_{\pm 0.61}$ & \textbf{47.08}$_{\pm 1.08}$ & +1.60 & 50.12$_{\pm 1.46}$ & \textbf{51.34}$_{\pm 1.09}$ & +1.22 \\
 & CAA      & 46.40 & \textbf{48.20} & +1.80 & 50.61 & \textbf{51.83} & +1.22 \\
 & Pause Token & 46.44$_{\pm 0.71}$ & \textbf{47.36}$_{\pm 1.68}$ & +0.92 & 50.98$_{\pm 0.33}$ & \textbf{52.07}$_{\pm 2.05}$ & +1.10 \\
 & Coconut  & 45.40$_{\pm 1.21}$ & \textbf{48.08}$_{\pm 1.32}$ & +2.68 & 52.07$_{\pm 0.70}$ & \textbf{52.56}$_{\pm 0.90}$ & +0.49 \\
\bottomrule
\end{tabular}
}
\end{table*}

We attribute this phenomenon to two key factors:
\begin{itemize}
    \item \textbf{Distribution Mismatch and Catastrophic Forgetting:} The official datasets contain ground-truth solutions that may not align with the base model's internal reasoning manifold. Forcing the model to mimic these "alien" distributions can disrupt its pre-trained knowledge, leading to catastrophic forgetting of its intrinsic capabilities. This is evident in the performance drop of LoRA, ReFT, and Coconut on MATH500 when trained on $\mathcal{D}_{full}$. In contrast, our $\mathcal{D}_{train}$ consists of \textit{self-generated} valid reasoning paths, ensuring that the training signal is strictly on-manifold.
    \item \textbf{Robustness of Soft CoT:} Interestingly, \textbf{Soft CoT} appears less susceptible to this negative transfer, likely because it learns a soft prompt to guide reasoning rather than modifying the model weights directly (or as extensively). However, it still benefits from the cleaner, aligned signal provided by our filtered data.
    \item \textbf{Data Efficiency and Coding Precision:} In the coding domain (HumanEval), our filtered dataset consistently outperforms or matches the full dataset. This suggests that for code generation, eliminating noise and ensuring correct, optimized solutions (as done in our pipeline) is more critical than raw volume.
    \item \textbf{Experimental Robustness:} We report the Mean $\pm$ Std over 5 independent runs for all experiments in Table~\ref{tab:data_scale_comparison}. The consistent trends across multiple models and methods, backed by low variance, strongly support our conclusions regarding the trade-offs between data scale and quality.
\end{itemize}

\section{Cross-Domain Generalization Analysis}
\label{sec:cross_domain}

To further evaluate the robustness of the learned reasoning patterns, we conducted a cross-domain generalization experiment. Specifically, we trained \our on the \textbf{MATH} dataset and evaluated it on the \textbf{HumanEval} coding task, and conversely, trained on the \textbf{Code} dataset and evaluated on \textbf{MATH500}. This setup tests whether the anchoring capabilities learned in one domain can transfer to another, indicating the acquisition of abstract, domain-agnostic reasoning strategies.

Table~\ref{tab:cross_domain} presents the results. We observe that models trained on mathematical reasoning demonstrate strong transfer performance to coding tasks, suggesting that the logical structuring learned from math problems is highly relevant to code generation. Similarly, models trained on code show competitive performance on math tasks, although the transfer is slightly less pronounced. This asymmetry might be due to the more rigid syntax requirements of code compared to the flexible reasoning paths in mathematics. Nevertheless, the positive transfer in both directions confirms that \our captures underlying reasoning manifolds that are shared across domains.

\begin{table}[h]
\centering
\small
\renewcommand{\arraystretch}{1.1}
\setlength{\tabcolsep}{3pt} 
\caption{\textbf{Cross-Domain Generalization.} Transferring reasoning patterns between Math and Code. $\mathcal{D}_{m}$ and $\mathcal{D}_{c}$ denote models trained on Math and Code data respectively. Zero-shot (ZS) represents the base model performance.}
\label{tab:cross_domain}
\resizebox{\columnwidth}{!}{
\begin{tabular}{lllc}
\toprule
\textbf{Model} & \textbf{Task} & \textbf{Source} & \textbf{Accuracy (\%)} \\
\midrule
\multirow{4}{*}{Qwen2.5-1.5B} 
& \multirow{2}{*}{HEval} & Zero-shot CoT & 46.34 \\
 & & Math ($\mathcal{D}_{m}$) & $\mathbf{53.41}_{\pm 0.70}$ \\
 \cmidrule(lr){2-4}
& \multirow{2}{*}{MATH} & Zero-shot CoT & 43.20 \\
 & & Code ($\mathcal{D}_{c}$) & $\mathbf{46.96}_{\pm 0.38}$ \\
\midrule
\multirow{4}{*}{Qwen2.5-7B} 
& \multirow{2}{*}{HEval} & Zero-shot CoT & 71.34 \\
 & & Math ($\mathcal{D}_{m}$) & $\mathbf{72.56}_{\pm 0.86}$ \\
 \cmidrule(lr){2-4}
& \multirow{2}{*}{MATH} & Zero-shot CoT & 71.00 \\
 & & Code ($\mathcal{D}_{c}$) & $\mathbf{73.84}_{\pm 0.41}$ \\
\bottomrule
\end{tabular}
}
\end{table}

\section{Implementation Details of \our}
\label{app:anon_details}

\subsection{HyperNetwork Architecture}
The HyperNetwork is implemented as a Multi-Layer Perceptron (MLP) that maps the context vector $\mathbf{c}$ to the affine transformation parameters $\gamma$ and $\beta$. Specifically, it consists of two linear layers with a hidden dimension equal to the model's hidden size $h$ (not $h/2$). The architecture is defined as:
\begin{equation}
    [\gamma, \beta] = \mathbf{W}_2 \cdot \text{Dropout}(\text{GELU}(\text{LayerNorm}(\mathbf{W}_1 \cdot \mathbf{c})))
\end{equation}
where $\mathbf{W}_1 \in \mathbb{R}^{h \times h}$ and $\mathbf{W}_2 \in \mathbb{R}^{2h \times h}$. The output is split into $\gamma \in \mathbb{R}^h$ and $\beta \in \mathbb{R}^h$. We apply Layer Normalization before the activation function to stabilize training.

\subsection{Normalization and Energy Scaling}
To ensure the injected anchor vector $\hat{\mathbf{z}}$ remains within the valid manifold of the base model, we apply a rigorous normalization and scaling process. The process follows this specific order:
\begin{enumerate}
    \item \textbf{Layer Normalization:} The raw output from the HyperNetwork is first normalized using LayerNorm.
    \item \textbf{L2 Normalization:} The vector is then projected onto the unit hypersphere via L2 normalization: $\mathbf{v} = \frac{\mathbf{v}'}{\|\mathbf{v}'\|_2}$.
    \item \textbf{Energy Scaling:} We re-scale the unit vector by the average energy of the context tokens ($\sigma_{ctx}$) to match the local activation magnitude.
    \item \textbf{Gate Scaling:} Finally, a learnable scalar gate $\alpha$ (initialized to 0) modulates the injection strength via a Softplus activation.
\end{enumerate}

\subsection{Layer Selection}
The choice of the insertion layer $l^\dagger$ is critical for effective anchoring. We empirically selected the insertion layers for different models and tasks as shown in Table~\ref{tab:insert_layers}. Generally, we target the deeper layers where abstract reasoning features are formed.

Concretely, both Qwen2.5-1.5B and Qwen2.5-7B contain 28 transformer layers, and we inject at $l^\dagger\!=\!26$ for Math to steer representations near the end of the computation pipeline, where the model aggregates long-range evidence and forms more abstract planning features. In contrast, for Programming we inject earlier (e.g., $l^\dagger\!=\!20$) to influence algorithmic structure and constraint satisfaction before the model commits to surface-form code tokens. For Llama-3.1-8B (32 layers), we follow the same principle by choosing a late layer for Math ($l^\dagger\!=\!31$) and a moderately earlier layer for Programming ($l^\dagger\!=\!25$), balancing sufficient depth for the anchor to propagate while avoiding overly late interventions that primarily affect decoding style rather than problem-solving trajectory.

\begin{table}[h]
\centering
\small
\caption{Insertion layer ($l^\dagger$) configuration for different models and tasks.}
\label{tab:insert_layers}
\begin{tabular}{lcc}
\toprule
\textbf{Model} & \textbf{Math Task} & \textbf{Programming Task} \\
\midrule
Qwen2.5-1.5B & Layer 26 & Layer 20 \\
Qwen2.5-7B   & Layer 26 & Layer 20 \\
Llama-3.1-8B & Layer 31 & Layer 25 \\
\bottomrule
\end{tabular}
\end{table}

\section{Implementation Details of Baselines}
\label{app:baselines}

We provide implementation specifications for the baseline methods used in our experiments. Training epochs are aligned with the main experiments (3 epochs for Math tasks, 10 epochs for Coding tasks) to ensure fair comparison.

\subsection{LoRA Baseline}
We utilize the standard Low-Rank Adaptation (LoRA) \citep{hu2022lora} as the primary baseline. We use the \texttt{peft} library to inject adapters into all linear projection layers of the attention mechanism. Key hyperparameters include Rank $r=16$, $\alpha=32$, and a learning rate of $1\text{e-}4$.

\subsection{ReFT Baseline}
We implement Representation Finetuning (ReFT) \citep{wu2024reft}, specifically the LoReFT variant. The intervention is modeled as a low-rank projection $h + R^T(Wh + b - Rh)$ applied to hidden states. We set the rank $r=16$ and use a higher learning rate of $4\text{e-}3$ to facilitate convergence of the intervention parameters.

\subsection{CAA Baseline}
We implement Contrastive Activation Addition (CAA) using the official open-source framework provided by \citet{panickssery2024steeringllama2contrastive}. Steering vectors are derived by averaging activation differences between successful guided reasoning paths and zero-shot failure cases on a held-out set. These static vectors are injected into the residual stream at all post-prompt positions. Following the official protocol, the steering coefficient $\alpha$ is optimized via grid search on the validation set. 

\subsection{Pause Token Baseline}
Following \citet{goyal2023think}, we insert learnable \texttt{<|pause|>} tokens to allow the model to utilize extra computation steps. During training, $N=8$ pause tokens are inserted between the prompt and the answer. The loss for these tokens is masked, allowing the model to autonomously learn their utility. We fine-tune both the embeddings and the LM head.

\subsection{Soft CoT Baseline}
We adopt the Soft Chain-of-Thought mechanism \citep{xu2025softcot}, and implement it based on the authors' official open-source codebase. It employs a small assistant model to generate "soft thoughts" for a larger base model. The hidden states from the assistant are projected to match the base model's dimension. We set the number of thought tokens to 10 and use a learning rate of $2\text{e-}5$, fine-tuning the assistant model alongside the projection layer.

\begin{figure*}[t]
\centering
\begin{minipage}[t]{0.48\textwidth}
\begin{tcolorbox}[
  colback=white, colframe=black, arc=3mm, width=\linewidth,
  title=\textbf{Case Study: Qwen2.5-7B-Instruct},
  coltitle=white, colbacktitle=gray, fonttitle=\bfseries
]
\scriptsize
\textbf{Prompt:}\\
You are an AI assistant; you are required to solve mathematical question step by step. Please provide your final answer on a separate line in \verb|\boxed{}| format.

\medskip

\textbf{Input:}\\
A Senate committee has 5 Democrats, 5 Republicans, and 1 Independent. In how many ways can they sit around a circular table if all the members of each party all sit next to each other? (Two seatings are considered equivalent if one is a rotation of the other.)

\medskip

\textbf{Baseline model output:}

To solve this problem, we need to consider both the arrangement of the groups (parties) around the table and the arrangement of individuals within each group.

\textbf{Step 1: Arrange the Groups}

Since the table is circular, fixing one group eliminates the rotational symmetry. We have three groups: Democrats (D), Republicans (R), and the Independent (I).
\begin{itemize}
  \item Fix one group in place. There are 2 remaining groups to arrange around the table.
  \item The number of ways to arrange 2 groups in a circle is given by \((2-1)!\).
\end{itemize}

\[
(2-1)! = 1! = 1
\]

So, there is only 1 way to arrange the groups around the table once one group is fixed.

\textbf{Step 2: Arrange Individuals Within Each Group}

\begin{itemize}
  \item \textbf{Democrats}: There are 5 Democrats, and they can be arranged among themselves in \(5!\) ways.
\[
5! = 120
\]
  \item \textbf{Republicans}: Similarly, there are 5 Republicans, and they can be arranged among themselves in \(5!\) ways.
\[
5! = 120
\]
  \item \textbf{Independent}: There is only 1 Independent, so there is only 1 way to arrange them.
\end{itemize}

\textbf{Step 3: Combine the Arrangements}

Now, we multiply the number of ways to arrange the groups by the number of ways to arrange the individuals within each group:

\[
1 \times 120 \times 120 \times 1 = 14400
\]

Thus, the total number of ways to seat the members around the table, considering that rotations are equivalent, is:

\[
\boxed{14400}
\]

\vspace{5pt}
\textbf{Model Answer:}

14400

\vspace{5pt}
\textbf{Evaluate:}

{\bf \color{red}Incorrect!}

\end{tcolorbox}
\end{minipage}
\hfill
\begin{minipage}[t]{0.48\textwidth}
\begin{tcolorbox}[
  colback=white, colframe=black, arc=3mm, width=\linewidth,
  title=\textbf{Case Study: Our Method },
  coltitle=white, colbacktitle=gray, fonttitle=\bfseries
]
\scriptsize
\textbf{Prompt:}\\
You are an AI assistant; you are required to solve mathematical question step by step. Please provide your final answer on a separate line in \verb|\boxed{}| format.

\medskip

\textbf{Input:}\\
A Senate committee has 5 Democrats, 5 Republicans, and 1 Independent. In how many ways can they sit around a circular table if all the members of each party all sit next to each other? (Two seatings are considered equivalent if one is a rotation of the other.)

\medskip

\textbf{Our model output:}

To solve this problem, we need to consider the arrangement of the groups (parties) and then the arrangement within each group.

\textbf{1. Arranging the Groups:}

Since the arrangement is circular and rotations of the same arrangement are considered identical, we fix one group in place to avoid counting rotations multiple times. We have 3 groups: Democrats (D), Republicans (R), and Independent (I).

Fixing one group, we have 2! ways to arrange the remaining 2 groups around the table. This is because the Independent can be placed in 2 relative positions to the fixed group (either to the left or right).

\textbf{2. Arranging Members Within Each Group:}

\begin{itemize}
  \item For the Democrats, there are 5 members, and they can be arranged in \(5!\) ways.
  \item For the Republicans, there are also 5 members, and they can be arranged in \(5!\) ways.
  \item The Independent has only 1 member, and there is only 1 way to arrange this member.
\end{itemize}

\textbf{3. Calculating the Total Number of Arrangements:}

The total number of ways to arrange the groups and the members within each group is the product of the number of ways to arrange the groups and the number of ways to arrange the members within each group.

Therefore, the total number of arrangements is:
\[
2! \times 5! \times 5! \times 1
\]

\textbf{4. Performing the Calculations:}

\begin{itemize}
  \item \(2! = 2\)
  \item \(5! = 120\)
\end{itemize}

Therefore, the total number of arrangements is:
\[
2 \times 120 \times 120 = 28800
\]

Thus, the number of ways the committee members can sit around the table, with all members of each party sitting next to each other, is \(\boxed{28800}\).

\medskip
\textbf{Final Answer:  \textbackslash boxed\{28800\}}

\vspace{5pt}
\textbf{Model Answer:}

28800

\vspace{5pt}
\textbf{Evaluate:}

{\bf \color{green}Correct!}

\end{tcolorbox}
\end{minipage}
\end{figure*}

\begin{figure*}[t]
\centering
\begin{minipage}[t]{0.48\textwidth}
\begin{tcolorbox}[
  colback=white, colframe=black, arc=3mm, width=\linewidth,
  title=\textbf{Case Study: Qwen2.5-1.5B-Instruct (Baseline, AIMO)},
  coltitle=white, colbacktitle=gray, fonttitle=\bfseries
]
\scriptsize
\textbf{Prompt:}\\
You are an AI assistant; you are required to solve mathematical question step by step. Please provide your final answer on a separate line in \verb|\boxed{}| format.

\medskip

\textbf{Input:}\\
Positive real numbers $x$ and $y$ satisfy $y^3=x^2$ and $(y-x)^2=4y^2$. What is $x+y$?

\medskip

\textbf{Baseline model output:}

To find the value of \( x + y \) given the equations \( y^3 = x^2 \) and \( (y - x)^2 = 4y^2 \), we will follow these steps:

1. \textbf{Express \( y \) in terms of \( x \) using the first equation:}
   \[
   y^3 = x^2 \implies y = x^{2/3}
   \]

2. \textbf{Substitute \( y = x^{2/3} \) into the second equation:}
   \[
   (y - x)^2 = 4y^2
   \]
   Substitute \( y = x^{2/3} \):
   \[
   \left( x^{2/3} - x \right)^2 = 4 \left( x^{2/3} \right)^2
   \]

3. \textbf{Simplify the equation:}
   \[
   \left( x^{2/3} - x \right)^2 = 4 x^{4/3}
   \]

4. \textbf{Take the square root of both sides:}
   \[
   x^{2/3} - x = \pm 2 x^{4/3}
   \]

5. \textbf{Check for integer solutions:}
   By trial and error with small integers, we find:
   \[
   x = 1 \implies y = 1^{2/3} = 1
   \]
   Check if it satisfies the original equations:
   \[
   y^3 = 1^3 = 1 \quad \text{(True)}
   \]
   \[
   (y - x)^2 = (1 - 1)^2 = 0 \quad \text{(True)}
   \]
   Therefore, \( x = 1 \) and \( y = 1 \).

6. \textbf{Calculate \( x + y \):}
   \[
   x + y = 1 + 1 = 2
   \]

Thus, the final answer is:
\[
\boxed{2}
\]

\vspace{5pt}
\textbf{Model Answer:}

2

\vspace{5pt}
\textbf{Evaluate:}

{\bf \color{red}Incorrect!}

\end{tcolorbox}
\end{minipage}
\hfill
\begin{minipage}[t]{0.48\textwidth}
\begin{tcolorbox}[
  colback=white, colframe=black, arc=3mm, width=\linewidth,
  title=\textbf{Case Study: Qwen2.5-1.5B-Instruct (Our Method, AIMO)},
  coltitle=white, colbacktitle=gray, fonttitle=\bfseries
]
\scriptsize
\textbf{Prompt:}\\
You are an AI assistant; you are required to solve mathematical question step by step. Please provide your final answer on a separate line in \verb|\boxed{}| format.

\medskip

\textbf{Input:}\\
Positive real numbers $x$ and $y$ satisfy $y^3=x^2$ and $(y-x)^2=4y^2$. What is $x+y$?

\medskip

\textbf{Our model output:}

To solve the given problem, we start with the equations:

1. \( y^3 = x^2 \)
2. \( (y - x)^2 = 4y^2 \)

First, let's expand and simplify the second equation:

\[
(y - x)^2 = 4y^2
\]

Expanding the left-hand side:

\[
y^2 - 2xy + x^2 = 4y^2
\]

Rearrange the equation to isolate terms involving \( y \):

\[
x^2 - 2xy = 3y^2
\]

Substitute \( x^2 = y^3 \) into the second equation:

\[
y^3 - 2xy = 3y^2
\]

Factor out \( y \) and divide by \( y \neq 0 \):

\[
y^2 - 3y = 2x
\]

Substitute \( x = \frac{y^2 - 3y}{2} \) into \( x^2 = y^3 \):

\[
\left( \frac{y^2 - 3y}{2} \right)^2 = y^3
\]

Simplify to get:

\[
y^2(y^2 - 10y + 9) = 0
\]

Since \( y > 0 \), solve \( y^2 - 10y + 9 = 0 \) giving \( y = 9 \) or \( y = 1 \). Checking the original constraints yields \( y=9 \) and \( x=27 \).

Finally,
\[
x + y = 27 + 9 = \boxed{36}
\]

\vspace{5pt}
\textbf{Model Answer:}

36

\vspace{5pt}
\textbf{Evaluate:}

{\bf \color{green}Correct!}

\end{tcolorbox}
\end{minipage}
\end{figure*}

\begin{figure*}[t]
\centering
\begin{minipage}[t]{0.48\textwidth}
\begin{tcolorbox}[
  colback=white, colframe=black, arc=3mm, width=\linewidth,
  title=\textbf{Case Study: Llama-3.1-8B-Instruct (Baseline)},
  coltitle=white, colbacktitle=gray, fonttitle=\bfseries
]
\scriptsize
\textbf{Prompt:}\\
You are a Python programmer. Output ONLY the function implementation. No explanations, no markdown, no comments.

\medskip

\textbf{Input:}\\
\texttt{HumanEval\_161\_solve}:\\
\texttt{def solve(s: str) -> str:}\\
\texttt{    \textbackslash"\textbackslash"\textbackslash You are given a string s.}\\
\texttt{    if s[i] is a letter, reverse its case from lower to upper or vise versa,}\\
\texttt{    otherwise keep it as it is.}\\
\texttt{    If the string contains no letters, reverse the string.}\\
\texttt{    The function should return the resulted string.}\\
\texttt{    Examples}\\
\texttt{    >>> solve('1234')}\\
\texttt{    '4321'}\\
\texttt{    >>> solve('ab')}\\
\texttt{    'AB'}\\
\texttt{    >>> solve('\#a@C')}\\
\texttt{    '\#A@c'}\\
\texttt{    \textbackslash"\textbackslash"\textbackslash }

\medskip

\textbf{Baseline model output:}
\begin{verbatim}
def solve(s: str) -> str:
    result = ''
    for char in s:
        if char.isalpha():
            result += char.swapcase()
        else:
            result += char
    return result[::-1]
\end{verbatim}

\vspace{5pt}
\textbf{Evaluate:}

{\bf \color{red}Incorrect!}

\end{tcolorbox}
\end{minipage}
\hfill
\begin{minipage}[t]{0.48\textwidth}
\begin{tcolorbox}[
  colback=white, colframe=black, arc=3mm, width=\linewidth,
  title=\textbf{Case Study: Llama-3.1-8B-Instruct (Our Method)},
  coltitle=white, colbacktitle=gray, fonttitle=\bfseries
]
\scriptsize
\textbf{Prompt:}\\
You are a Python programmer. Output ONLY the function implementation. No explanations, no markdown, no comments.

\medskip

\textbf{Input:}\\
\texttt{HumanEval\_161\_solve}:\\
\texttt{def solve(s: str) -> str:}\\
\texttt{    \textbackslash"\textbackslash"\textbackslash You are given a string s.}\\
\texttt{    if s[i] is a letter, reverse its case from lower to upper or vise versa,}\\
\texttt{    otherwise keep it as it is.}\\
\texttt{    If the string contains no letters, reverse the string.}\\
\texttt{    The function should return the resulted string.}\\
\texttt{    Examples}\\
\texttt{    >>> solve('1234')}\\
\texttt{    '4321'}\\
\texttt{    >>> solve('ab')}\\
\texttt{    'AB'}\\
\texttt{    >>> solve('\#a@C')}\\
\texttt{    '\#A@c'}\\
\texttt{    \textbackslash"\textbackslash"\textbackslash }

\medskip

\textbf{Our model output:}
\begin{verbatim}
def solve(s: str) -> str:
    result = []
    for char in s:
        if char.isalpha():
            result.append(char.swapcase())
        else:
            result.append(char)
    if not any(char.isalpha() for char in s):
        return s[::-1]
    return ''.join(result)
\end{verbatim}

\vspace{5pt}
\textbf{Evaluate:}

{\bf \color{green}Correct!}

\end{tcolorbox}
\end{minipage}
\end{figure*}